\documentclass[12pt]{article} 

\usepackage[paperheight=9.44in,paperwidth=6.69in,left=1.85cm,right=1.8cm,bottom=2.5cm,top=3cm,twoside]{geometry}

\usepackage{subfig} 

\usepackage[ruled,vlined,boxed]{algorithm2e}
\usepackage{authblk} 
\usepackage{fancyhdr} 
\usepackage{lastpage} 
\usepackage{textcomp} 
\usepackage{amsmath} 
\usepackage{amssymb} 
\usepackage{amsthm} 
\usepackage{graphicx} 
\usepackage{multirow} 
\usepackage{hyperref} 
\usepackage{longtable} 
\usepackage{float} 
\usepackage{enumitem} 
\usepackage[T1]{fontenc} 
\usepackage[utf8]{inputenc} 

\newcommand{\linia}{\noindent\rule{\linewidth}{0.25mm}\hrulefill} 

\SetAlFnt{\small}
\SetAlCapFnt{\small}
\SetAlCapNameFnt{\small}
\SetAlCapHSkip{0pt}
\IncMargin{-\parindent}

\usepackage{times}
\usepackage{titlesec}

\titleformat*{\section}{\large\bfseries}
\titleformat*{\subsection}{\normalsize\bfseries}

	\fancypagestyle{firststyle}
	{
		\fancyhf{}
		\setcounter{page}{41} 
		\fancyfoot{}
		\fancyhead{}
		\fancyhead[CE]{\upshape Zeszyty Naukowe WWSI, No 31, Vol. 19, 2024, pp. \thepage-\pageref{LastPage} \newline DOI: 10.26348/znwwsi.31.\thepage}
		\fancyhead[CO]{\upshape Zeszyty Naukowe WWSI, No 31, Vol. 19, 2024, pp. \thepage-\pageref{LastPage} \newline DOI: 10.26348/znwwsi.31.\thepage}
		\fancyfoot[L,R]{\footnotesize\bfseries Manuscript received October 11, 2024}
		\fancyfoot[LE,RO]{\thepage}
		}

\title{\large \bfseries New Approach to Clustering Random Attributes} 
\author{\normalsize Zenon Gniazdowski\thanks{E-mail: zgniazdowski@wwsi.edu.pl}}
\affil{\normalsize Warsaw School of Computer Science}

\date{\vspace{-5ex}}
\providecommand{\keywords}[1]{\textbf{\textit{Keywords ---}} #1}

\begin{document}

\maketitle 
\thispagestyle{firststyle} 

\linia

\begin{abstract}\label{abstract}
\noindent This paper proposes a new method for similarity analysis and, consequently, a new algorithm for clustering different types of random attributes, both numerical and nominal. However, in order for nominal attributes to be clustered, their values must be properly encoded. In the encoding process, nominal attributes obtain a new representation in numerical form. Only the numeric attributes can be subjected to factor analysis, which allows them to be clustered in terms of their similarity to factors.  The proposed method was tested for several sample datasets. It was found that the proposed method is universal. On the one hand, the method allows clustering of numerical attributes. On the other hand, it provides the ability to cluster nominal attributes. It also allows simultaneous clustering of numerical attributes and numerically encoded nominal attributes.
\end{abstract}
\keywords{\small numerical attributes, nominal attributes, numerical encoding, similarity of random attributes, clustering of random attributes, factor analysis}\label{keywords}

\section{Introduction}\label{intro}
The goal of data analysis is to discover the hidden structure of data. One way to discover the structure of data is to cluster it. Clustering can be talked about in two different contexts.  On the one hand, it is the clustering of points (objects) belonging to a certain space defined by random attributes. In a given space, clusters are identified in such a way that within a cluster the points are maximally similar to each other, while between clusters there should be maximal dissimilarity. On the other hand, random attributes can also be clustered. Two random attributes are similar if they are correlated. Applying factor analysis to correlated random attributes, it is also possible to cluster these attributes due to their similarity to factors.

Both the clustering of points and the clustering of random attributes require operating on numbers: the coordinates of the points are numbers, but also the values taken by the random attributes are also numbers. In number spaces, it is possible to measure the dissimilarity of points, defined as their distance, but it is also possible to measure the similarity of random attributes, defined as a coefficient of determination measuring their common variance.

On the other hand, data analysis examines data measured using different measurement scales: ratio scale, interval scale, ordinal scale and nominal scale \cite{Stevens1946}. Different types of data require different specific methods of analysis. However, particular difficulties arise when numerical and nominal data are analyzed simultaneously. Two approaches are then possible:
\begin{itemize}[nosep]
\item In the classical approach, data measured at a stronger scale are transformed so that they can be interpreted as data measured at a weaker scale. For example, when numerical and nominal data are analyzed simultaneously, then all numerical data belonging to a certain assumed range of values are treated as one separate categorized value, which is interpreted as a nominal value. This way leads to the loss of many natural, useful features of numerical data. First of all, the possibility of ordering the data is given up here. And since nominal data can only be counted, therefore statistics such as, for example, the $\chi^2$ statistic or Cramer's $V^2$ coefficient can be calculated from the contingency table. Subsequently, these statistics can be used to test the significance of a correlation relationship or to assess the strength of a correlation relationship \cite{Blalock1960}. 
\item There is also another approach in which nominal data are encoded using numbers. Such encoding not only does not lose any information about the nominal random attribute, but also enriches the knowledge about it with additional, previously unknown facts \cite{Gniazdowski2015}.
\end{itemize}
This article will propose a new algorithm for clustering random attributes, whether they are numerical random attributes or nominal random attributes. For nominal random attributes, this new algorithm will use encoding of nominal data with numbers.  After encoding, nominal random attributes will receive a new representation in numerical form. 
In the case of numerical random attributes, it is possible to calculate their mutual correlations. Then factor analysis can be used to analyze the similarities between these random attributes and factors.
This approach will make it possible to cluster not only random attributes with numerical values or numerically encoded nominal random attributes, but also to cluster both types of random attributes simultaneously. 
The clustering algorithm will implement factor analysis for the numerical representation of the values of random attributes, both in pure form and in ranked form. Factor analysis will be used to model the numerical representation of the above random attributes as a function of factors. 
The number of factors should be chosen so that the factor model reconstructs most (i.e., more than half) of the variance of all modeled random attributes. 
In this article, it is arbitrarily assumed that enough factors should be selected so that the factor model reconstructs at least $55\%$ of the variance for each modeled random attribute. 
At the input of the final clustering procedure will be given the matrix of common variances calculated for the final factor model obtained after Varimax rotation.

The proposed algorithm will be tested for several datasets. In the first example, the successive steps of the described algorithm will be shown in detail. 
So that this can be done, the data -- by discarding some of its properties -- will be treated as a useful dummy. In later examples, detailed descriptions of the steps of the algorithm will be omitted, and instead emphasis will be placed on presenting the final result of the algorithm as a directed graph. Since the purpose of the article is not to analyze the data, but only to demonstrate the new algorithm, therefore, after the graphical presentation of the result of the algorithm, the author will omit the detailed analysis of this result, leaving it as the subject of possible further work.

\section{Preliminaries}
Since the proposed method is based on previously known facts and methods, therefore, the preliminaries will signal these useful facts and methods relating to both  nominal data and numerical data.
\subsection{Numerical measurement scale vs. nominal measurement scale}\label{numVSnom}
Data analysis examines different types of data, measured using different measurement scales \cite{Stevens1946}. Data are measured on ratio, interval and ordinal scales, as well as on a nominal scale. The ratio scale and interval scale are numerical scales. Data measured on an ordinal scale can be numbers or can obtain a numerical representation, even if they are not originally numbers. 
The latter situation, for example, can occur with data measured on a Likert scale, when the data can take the following values: $Strongly~Disagree$e, $Disagree$, $Neutral$, $Agree$, $Strongly~Agree$. These values can be sorted according to their strength, from weakest to strongest. It is possible to assign numerical equivalents to values that have been sorted:
$Strongly~Disagree=-2$, $Disagree=-1$, $Neutral=0$, $Agree=1$, $Strongly~Agree=2$.  In the next step, ranks can also be assigned to them \cite{Berry2018}.

In principle, data can be distinguished based on the type of mathematical relation that can be defined in a given data set. On the one hand, the linear (total) order relation can be defined in sets measured by ratio scale, interval scale, as well as by ordinal scale. This means that two different elements in a set are equivalent, or one element precedes the other. On the other hand, in a set containing data measured on a nominal scale, a linear order relation cannot be defined, but only an equivalence relation can be defined. That is, it means that two elements in a set can be either equivalent or different, but no element can precede another element.
This fundamental difference regarding the mathematical relations that can be defined in sets measured at different measurement scales leads to the conclusion that data can be divided into two classes. Such a division will be made for the purposes of this paper. The remainder of the paper will refer to numerical data and nominal data. The class of numerical data will include data measured on ratio scales, interval scales, as well as ordinal scales. 

\subsection{Attribute and class}
By attribute is usually meant a characteristic feature of some object that can be measured. If the measured values of an attribute take on random values, then the attribute can be thought of as a random variable. For this reason, both random variable and attribute are sometimes equated. In the literature, the two terms are used interchangeably \cite{Hastie2017}\cite{Han2022}.

The type of an attribute (random variable) is determined by the set of possible values that the attribute can take. These can be nominal or numeric values. A numeric attribute is a random variable that takes numeric values that can be ordered (sorted). A nominal attribute is a random variable that takes different nominal values relating to names. Nominal values cannot be ordered (sorted) \cite{Stevens1946}\cite{Han2022}.
It can be noted that a nominal attribute (nominal random variable) can be viewed as a set consisting of classes of equivalent elements. 
That is, within a given attribute, each class is a subset that contains all of the elements that have identical nominal values.

This article focuses on the clustering of nominal attributes, which are described as random variables. Since both random variable and attribute are sometimes equated in the literature, therefore in the remainder of this article the two terms will also be used interchangeably.
The random variable will be discussed more frequently in the section entitled ''Preliminaries'', where the necessary theoretical basis of issues concerning numerical random variables will be presented. Here we will not refer to concretized random variables, but to abstract concepts.
On the other hand, the concept of attribute will be used in later sections of this article, when the description and study of specific numerical random variables, as well as nominal random variables, both uncoded and numerically encoded, will be presented.

\subsection{The random component of a numerical random variable as a vector}
A numerical random variable $X$ is considered. In particular, a random sample of size $n$ is available. The elements $X_i$ represent the $i$-th realization of the random variable:
\begin{equation}
X=\{X_1,X_2,\ldots,X_n\}.
\end{equation}
The estimator of the expected value of a random variable $X$ is its average value:
\begin{equation}
\overline{X}=\frac{1}{n}\sum_{i=1}^{n}X_i.
\end{equation}
In order to be able to assess the dispersion of the variable $X$, it is necessary to subtract the average value of $X$ from each individual realizations of the $X$ variable. This will give the random component $x$ of the random variable $X$.
The individual realizations of the $x$ variable will take the form:
\begin{equation}\label{rndCmp}
x_i=X_i-\overline{X}.
\end{equation}
This is a random variable $X$ reduced by a constant component. The random component $x$ is a vector in n-dimensional space:
\begin{equation}
x=\{x_1,x_2,\ldots,x_n\}.
\end{equation}
\subsection{The effect of a linear transformation of a numerical random variable on the direction of its random vector}\label{lin2vect}
A numerical random variable $X$ can be represented as the sum of its average value $\overline{X}$ and its random component $x$:
\begin{equation}
X=\overline{X}+x.
\end{equation}
It can be seen that the linear transformation ($a+bX$) of the random variable $X$ does not change the direction of the vector representing its random component:
\begin{equation}
a+bX=a+b(\overline{X}+x)=a+b\overline{X}+bx.
\end{equation}
After a linear transformation, the constant $a+b\overline{X}$ represents the average value of the transformed variable, and the random vector $x$ becomes the new random component $bx$ of the transformed variable. The vector $bx$ is parallel to the vector $x$. This means that a linear transformation of a random variable does not change the direction of its random vector \cite{Gniazdowski2022}. 
\subsection{Correlation between two numerical random variables as the cosine of the angle between their random vectors}
A measure of the relationship between two numerical random variables $X$ and $Y$ is their covariance.
Covariance normalized to unity is called the Pearson correlation coefficient:
\begin{equation}
R_{X,Y}=\frac{\sum_{i=1}^{n}\left[\left(X_i-\overline{X}\right)\left(Y_i-\overline{Y}\right)\right]}{\sqrt{\sum_{i=1}^{n}\left(X_i-\overline{X}\right)^2}\sqrt{\sum_{i=1}^{n}\left(Y_i-\overline{Y}\right)^2}}.
\end{equation}
Using (\ref{rndCmp}), the formula for the correlation coefficient can be transformed to the form:
\begin{equation}
R_{X,Y}=\frac{\sum_{i=1}^{n}{x_iy_i}}{\sqrt{\sum_{i=1}^{n}x_i^2}\sqrt{\sum_{i=1}^{n}y_i^2}}.
\end{equation}
The numerator of the above formula for the correlation coefficient contains the scalar product of the two vectors $x$ and $y$, while the denominator contains the product of the lengths of these vectors. This means that the correlation coefficient is identical to the cosine of the angle between the two random vectors $x$ and $y$ \cite{Gniazdowski2013}:
\begin{equation}\label{cosine}
R_{X,Y}=\frac{\sum_{i=1}^{n}{x_iy_i}}{\sqrt{\sum_{i=1}^{n}x_i^2}\sqrt{\sum_{i=1}^{n}y_i^2}}=\frac{x\cdot y}{\|x\|\cdot\|y\|}=cos(x,y).
\end{equation}
\subsection{Linear transformation of a numerical random variable vs. correlation coefficient}\label{lin2corel}
A linear transformation of a numerical random variable does not change the direction of a random vector, but at most can change its orientation along a given direction \cite{Gniazdowski2022}. For this reason, a linear transformation of a numerical random variable does not affect the value of the cosine of the angle between the random vectors.
On the other hand, by changing the orientation of the random vector, the transformation of the random variable can affect the sign of the scalar product in formula (\ref{cosine}). This means that the linear transformation of the random variable can only affect the sign of the cosine of the angle between the random vectors. 

Since the correlation coefficient has the interpretation of the cosine of the angle between the random components of the random variable, the linear transformation of the random variable does not affect the value of the correlation coefficient. 
A linear transformation of a random variable can at most change the sign of this correlation coefficient. 
Both standardization and normalization of a numerical random variable are linear transformations that change neither the direction of the random vector nor its orientation along a given direction. Thus, both standardization and normalization of a numerical random variable affect neither the value nor the sign of the correlation coefficient.

\begin{table}[h!]
\centering
\caption{Ranking of a numerical set}\label{rankSet}
\fontsize{8.5}{12}\selectfont{
	\begin{tabular}{|c|c|c|}
		\hline
		No.&Sorted data&Assigned ranks\\ \hline
		1&21&1\\
		2&28&2\\
		3&33&3\\
		4&44&4\\
		5&45&5\\
		6&54&6\\
		7&55&7\\
		8&60&8\\
		9&63&9\\
		10&76&10\\ \hline
	\end{tabular}
}
\end{table}

\begin{table}[h!]
\centering
\caption{Ranked set with tied ranks}\label{tied}
\fontsize{8.5}{12}\selectfont{
	\begin{tabular}{|c|c|c|}
		\hline
		No.&Sorted data&Assigned ranks\\ \hline
		1&21&1\\
		2&28&2\\
		3&44&4\\
		4&44&4\\
		5&44&4\\
		6&54&6\\
		7&55&8.5\\
		8&55&8.5\\
		9&55&8.5\\
		10&55&8.5\\ \hline
	\end{tabular}
}
\end{table}

\subsection{Ranking of a random variable}
Ranking of numerical data involves replacing observations with ranks \cite{Wilcoxon1992}. The ranking procedure consists of two stages. In the first stage, the data set is sorted non-decreasingly. In the second stage, successive elements from the sorted set are assigned a rank equal to the position of the element in the sorted set. After the rank assignment, further operations are not performed on the original variables, but on the ranks. An example of rank assignment is shown in Table \ref{rankSet}.

If there are numbers with identical values in the sorted set, then a single rank is assigned for all identical values. This is known as the tied rank. The tied rank is equal to the average position of all equal elements in the sorted set. Table \ref{tied} shows an example of a set with tied ranks.

Although ranking is a transformation of a random variable, it is not a linear transformation. For this reason, vectors containing random components calculated for data before and after ranking may have different directions. Consequently, the ranking may affect the value of the correlation coefficient.
On the other hand, if a numerical random variable is a dichotomous variable, i.e. a variable that takes only two values, then its rank is always a linear transformation. Therefore, normalization or standardization of dichotomous numerical random variables does not affect the values of calculated correlation coefficients\footnote{The author is aware that standardization can only be implemented for random variables with a normal distribution. In the case of a dichotomous variable, its standardization can only be considered as a formal procedure.}.
\subsection{Numerical encoding of nominal data}\label{numCod}
In ranking, the value of assigned ranks depends on two factors. On the one hand, the rank value depends on the order of the element in the sorted set. On the other hand, in the case of the appearance of identical elements, the value of rank also depends on the cardinality of these elements. Meanwhile, in a set containing nominal values, it is not possible to establish order relations, and consequently nominal values cannot be sorted. As a result, it is not possible to talk about the order of elements in the set. This means that a possible attempt to assign ranks to nominal elements cannot be based on their values.

Also, intuition tells us that in a statistical set, whether numerical or nominal, an element occurring more frequently is more important than an element with less cardinality. Therefore, the idea was born to rank the different nominal values only on the basis of their cardinality \cite{Gniazdowski2015}.
\subsubsection{The case of different cardinalities of nominal values}
In a given n-element subset of identical nominal values, elements can be numbered with numbers from 1 to n. Each element can be assigned a rank equal to the average value from the assigned numbers:
\begin{equation}\label{ranga}
R=\frac{\left(n+1\right)}{2}
\end{equation}
The rank of the elements from the subset with greater cardinality will be greater than the rank of the elements from the set of lesser cardinality. Thus, a linear order can be introduced in the set of nominal values.
\subsubsection{The case of equal cardinalities of different nominal values}
When there are different elements with identical cardinalities in a set containing nominal values, the method described above would give identical ranks for different values with equal cardinalities. This would lead to a situation in which different elements with equal cardinalities would be indistinguishable. For this reason, the method should be modified. If a nominal variable has $k$ different values with identical cardinalities, then the elements of the $j$-th subset (j=0,1,\ldots,k-1) can be encoded by the $k$-th degree roots of unity:
\begin{equation}
R_j=R\cdot\sqrt[k]{1}=R\cdot e^{i\varphi_j}.
\end{equation}
In the above expression, $i=\sqrt{-1}$ is the imaginary unit, $\varphi_j=\frac{2\pi j}{k}$ is the phase assigned to the $j$-th nominal value, $R$ is the complex rank modulus depending on the cardinality of the $j$-th nominal value, calculated using formula (\ref{ranga}). This approach gives identical rank modules for values with identical cardinalities, while it distinguishes their ranks using different phases.
\subsubsection{Properties of numerical encoding of nominal data}
When a nominal random variable is encoded with numbers, the encoded data obtains additional properties:
\begin{itemize}[nosep]
\item The value of the $R$-module contains information about the cardinality of a given subset of identical elements, and therefore contains information about the statistical strength of this subset.
\item The different values of $\varphi$ phases accompanying a given R-module result from the number of equicardinal classes of different values in the set.
For this reason, the $\varphi$ phase contains information about the number of equicardinal classes of different values of the random variable.
\item Points in the space of nominal data acquire the properties of vectors in a numerical space (real or complex). In such a space, it is possible to define a scalar product and, consequently, a metric. Thus, numerically encoded nominal data can be clustered, as well as classified. 
\end{itemize}
\subsubsection{Correlations for numerically encoded nominal data}\label{cod2cor}
The correlation coefficient can only be estimated for real random variables, that is, variables that have been measured on a ratio scale, interval scale or ordinal scale. A common feature of data sets measured on the above three measurement scales is that these sets can be sorted, and consequently they can also be assigned ranks\footnote{For ranks, the correlation coefficient can be estimated. This is known as the Spearman rank correlation coefficient \cite{Blalock1960}.}. Thus, these are sets for which a linear order relation can be established. 
A necessary condition for estimating an interpretable correlation coefficient between two random sets is the possibility of the existence of total order relations in the sets containing both random variables \cite{Gniazdowski2022}.

If a random variable with nominal values does not have classes with identical cardinalities, then this variable can be encoded with real numbers. Since a linear order relation can be established in the set of real numbers, so the random variable encoded in this way can be analyzed for correlation relationships.

On the other hand, for sets measured on a nominal scale, having equicardinal classes of identical elements encoded by complex numbers, a linear order relation cannot be defined. In a set of complex numbers, the strongest definable relation is the partial order relation. Therefore, for nominal data containing classes with identical cardinalities, it is not possible to estimate an unambiguously interpretable correlation coefficient \cite{Gniazdowski2022}. In such a situation, complex encoding should be abandoned, and other possible solutions should be considered instead.
\subsection{Factor analysis}\label{FA}
Based on a set of observed correlated random variables, linear models of these variables are built as functions of a set of factors that are independent random variables with unit variance. The factors are subject to interpretation. If the factors are interpreted, the causes of variation in the observed random variables can be inferred.

In this article, only exploratory factor analysis based on principal components, which uses the Varimax method to rotate factors, will be discussed and used. 
In the factor analysis discussed here, three stages can be distinguished. 
First, a matrix $L$ containing the complete set of factor loadings is calculated. In the next stage, the $L$ matrix is reduced. The reduction of the $L$ matrix is equivalent to selecting a smaller number of factors that are sufficient to reproduce most of the variance of the primary variables being modeled. Finally, the factors are rotated so that a given modeled primary variable is more similar to a single factor and little similar to other factors. 

\begin{algorithm}[h!]
\centering
\caption{Algorithm for calculating the full factor loading matrix}\label{algorFulFact}
\fontsize{10}{12}\selectfont{
	\begin{enumerate}[nosep]
		\item Solve the eigenproblem for the correlation coefficient matrix $R$. Obtain the eigenvalues and eigenvectors:
		\begin{enumerate}[nosep]
			\item 	From the non-increasingly sorted eigenvalues of $\lambda_i$, create a diagonal matrix $\Lambda$:
			\begin{equation}\label{r24}
				\Lambda=\left[\begin{matrix}\lambda_1&\cdots&0\\\vdots&\ddots&\vdots\\0&\cdots&\lambda_n\\\end{matrix}\right].
			\end{equation}
			\item From successive eigenvectors corresponding to successive eigenvalues, create successive columns of matrix $U$:
			\begin{equation}\label{r25}
				U=\left[\begin{matrix}U_{11}&\cdots&U_{1n}\\
					\vdots&\ddots&\vdots\\U_{n1}&\cdots&U_{nn}\\
				\end{matrix}\right].
			\end{equation}
		\end{enumerate}
		\item By rooting the diagonal matrix $\Lambda$, create the diagonal matrix $S$:
		\begin{equation}\label{r26}
			S=\sqrt\mathrm{\Lambda}=\left[\begin{matrix}\sqrt{\lambda_1}&\cdots&0\\
				\vdots&\ddots&\vdots\\0&\cdots&\sqrt{\lambda_n}\\
			\end{matrix}\right].
		\end{equation}
		\item From the product of the matrix $U$ and the matrix $S$, create a matrix $L$ that contains the factor loadings.
		\begin{equation}\label{r27}
			L=U\cdot S=\left[\begin{matrix}L_{11}&\cdots&L_{1n}\\
				\vdots&\ddots&\vdots\\L_{n1}&\cdots&L_{nn}\\
			\end{matrix}\right].
		\end{equation}
	\end{enumerate}
}
\end{algorithm}

\subsubsection{Calculation of the full factor loading matrix}
The starting point for factor analysis is a matrix containing $n$ random variables $X=\left[X_1,\ldots,X_n\right]$. Successive $X_i$ variables are successive columns of the $X$ matrix. A matrix of correlation coefficients $R$ is calculated for the matrix $X$.
If there is a calculated $R$ matrix that contains correlation coefficients, the algorithm for calculating the full factor loading matrix (Algorithm \ref{algorFulFact}) can be implemented in several steps \cite{Gniazdowski2021}.

\begin{algorithm}[h!]
\centering
\caption{ The algorithm for selection of factors }\label{algor1}
\fontsize{10}{12}\selectfont{
	\begin{tabular}{r|rcl|c}
		Input:&$L$&--&\multicolumn{2}{l}{Full matrix of factor loadings}  \\
		&$\varepsilon$&--&\multicolumn{2}{l}{minimum acceptable level of representation }\\
		&&&\multicolumn{2}{l}{of the variance of a single  variable -- assume $\varepsilon>0.5$}  \\  
		Output:&NoF&--&\multicolumn{2}{l}{Number of factors}  \\
		Notes:&$V$&--&\multicolumn{2}{l}{Matrix of common variances}\\
		&$VC$&--&\multicolumn{2}{l}{Matrix of cumulative common variances}\\
		\hline 
		Step No.&&&Steps of the algorithm & Comments\\ \hline 
		1.&&&for (i=0; i<n; i++)&Hadamard product:  \\
		&&&\quad for (j=0; j<n; j++)&$\forall_{i,j}V_{ij}=L_{ij}^2 $ \\
		&&&\quad\quad $V_{ij} = L_{ij}^2$;&  \\ \hline
		2.&&&for (i=0; i<n; i++)&The first column of matrix $VC$:  \\
		&&&\quad $VC_{i0} = V_{i0};$&$\forall_iVC_{i0}=V_{i0}$  \\ \hline
		3.&&&for (j=1; j<n; j++)&Successive columns of matrix $VC$:  \\
		&&&\quad for(i=0; i<n; i++)&$\forall_{j>0}(\forall iVC_{ij}=VC_{ij-1}+V_{ij})$  \\
		&&&\quad\quad $VC_{ij} = VC_{ij-1} + V_{ij}$;&\\ \hline
		4.&&&for (j=0; j<n; j++)&Minimum in the $j$-th column   \\
		&&&\{\,  $minVar = VC_{0j}$;&  of matrix $VC$\\ 
		&&&\quad  for (i=1; i<n; i++)&  \\ 
		&&&\quad\quad  if ($VC_{ij} < minVar$)&  \\ 
		&&&\quad\quad\quad  $minVar = VC_{ij}$;&  \\ 
		&&&\quad  if ($minVar > \varepsilon$)&  \\ 
		&&&\quad\quad  break;&  \\ 
		&&&\}&  \\  \hline
		5.&&&NoF = j+1;&Selected number of factors  \\ \hline
		6.&&&return NoF;&  
	\end{tabular}
}
\end{algorithm}

\subsubsection{Selecting the right number of factors}
Since the factors $f_1,\ldots,f_n$ are standard independent random variables, therefore the square of a given factor loading $L_{ij}$ represents the variance contributed by a given factor $f_j$ to a given random variable $X_i$. Since the eigenvalues in the matrix $\Lambda$ are ordered non-increasingly, therefore the effect of the first $k$ factors on the $X_i$ variables is dominant over the last $n-k$ factors.
Therefore, the $X_{i}$ variables can be influenced by the $k$ first factors, ignoring the $n-k$ last factors. For this purpose, the square matrix $L_{n\times n}$ is replaced by a rectangular matrix $L_{n\times k}$.
Usually several methods are used to select the right number of factors:
\begin{itemize}[nosep]
\item Percentage criterion of the part of the variance explained by factors -- as many factors are taken so that the sum of the eigenvalues associated with successive factors related to the trace of the matrix of correlation coefficients is greater than the assumed minimum threshold.
\item Scree plot criterion -- as many factors as many eigenvalues are taken on the scree slope.
\item Eigenvalue criterion -- the number of factors is equal to the number of eigenvalues not less than 1.
\item The criterion of the number of factors -- the number of factors should be less than half the number of variables modeled. 
\end{itemize}
None of these methods is universal. The last two methods are the simplest, but represent only intuitively obvious hopes that are sometimes unreliable in practical applications. The scree criterion sometimes cannot be applied because the graph does not show its breakdown in the form of the so-called elbow. The first method leads to a model that represents some assumed level of variance of the modeled variables, but it is only an average level \cite{Gniazdowski2021}. This method leads to a factorial model that can reproduce the assumed average variance of the standardized modeled primary variables.
Unfortunately, an adequate representation of the average variance of the modeled variables does not mean that the factor model represents most of the variance of the individual modeled variables. Despite the reproduced acceptable average variance for all modeled variables, the variance of some individual variables may be underrepresented. This may mean that the representation of some variables does not reproduce even half of their variance.

The answer to the problems that come from using the above four methods is the method described in the paper \cite{Gniazdowski2021}. This method creates a matrix $V$ from the loadings matrix $L$, which contains the common variances between the variables modeled and the factors. The $V$ matrix is the Hadamard square of the $L$ matrix \cite{Styan1973}:
\begin{equation}
\forall_{i,j}V_{ij}=L_{ij}^2.
\end{equation}
The element $V_{ij}$ tells what part of the variance of the $i$-th modeled variable is represented by the $j$-th factor. A given $i$-th modeled variable is reproduced by summing successive factors appropriately weighted by factor loadings.
Since the factors are independent, the variance of the modeled variable that is reconstructed by the successive factors is equal to the sum of the partial variances contributed by these factors.
Based on the variance matrix $V$, the cumulative variance matrix $VC$ can be calculated. In the $VC$ matrix, the successive columns contain the variance levels of the modeled variables, which are reconstructed by more and more successive factors. Thus:
\begin{itemize}[nosep]
\item In the first column of the matrix $VC$, the successive rows contain that part of the variance of each modeled variable that is reproduced by the first factor. The average value of the elements in the first column will be equal to the average variance of all modeled variables reproduced by the first factor. The average value of the elements in the first column is identical to the cumulative percentage of explained variance in the percentage criterion of the part of the variance explained by a single factor.
\item In the second column of the matrix $VC$, the successive rows contain that part of the variance of each modeled variable that is reproduced by the first two factors. The average value of the elements in the second column will be equal to the average variance of all modeled variables reproduced by the first two factors. The average value of the elements in the second column is identical to the cumulative percentage of explained variance in the percentage criterion of the part of the variance explained by the two factors.
\item Successive columns of the $VC$ matrix will contain information relating to the impact of successive factors on the level of reproduction of subsequent modeled variables. The averages in the successive columns are identical to the cumulative percentages of explained variance in the percentage criterion of the part of the variance explained by the successive factors.
\end{itemize}
The average values in each column of the $VC$ matrix correspond to the percentage criterion of the part of the variance explained by the factors. 
It may happen that in the $k$-th column of the $VC$ matrix, the average value of the variance is satisfactory from the point of view of the percentage criterion of the part of the variance explained by the factors. This means that it can be assumed that $k$ factors are enough to model the variables.
On the other hand, in the same $k$-th column of the $VC$ matrix, in a certain row, there may be a value that is unsatisfactory from the point of view of the representation of a certain modeled random variable. In this case, it is not possible to rely on the percentage criterion of the part of the variance explained by the factors.

The algorithm for finding the right number of factors should identify enough factors to represent each of the modeled variables at least satisfactorily. This means that the selected factors should represent most of the variance of each variable, that is, more than half of the variance of each variable. In this work, it is assumed that the minimum acceptable level of reproduced variance is not less than $55\%$. 

To find the appropriate number of factors, Algorithm \ref{algor1} is used, which finds the minimum value in the successive columns of the VC matrix. If in a given column the minimum value found is less than the acceptable variance threshold, then the algorithm moves on to the next column. The procedure ends when the minimum value found in a given column is greater than the assumed minimum threshold of explained variance. The number of the column in which the search ended is also the minimum number of factors.

The procedure for selecting the appropriate number of factors led to a situation where the last few columns of the $L$ matrix were rejected. The square matrix $L_{n\times n}$ was replaced by a smaller matrix $L_{n\times k}$, which contains only $k$ columns $(k<n)$. This means that from now on only $k$ factors will be used to model the standardized primary variables. 
In practice, this means that for a given random point  $f=[f_1,f_2,\ldots,f_k]$ in the $k$-dimensional space of independent factors, it is possible to simulate a point $x=[x_1,x_2,\ldots,x_n]$ in the $n$-dimensional space of standardized primary variables:
\begin{equation}
x^T=L\cdot f^T.
\end{equation}
The above notation, which is suitable for a single point, can be generalized to matrix notation. 
If given a random matrix  $F_{m\times k}$ that represents $m$ points in a $k$-dimensional space of independent standardized random variables called factors, then using this matrix it is possible to represent (simply simulate) a matrix $X_{m\times n}$ that contains $m$ points in an $n$-dimensional space of standardized primary variables:
\begin{equation}
X^T=L\cdot F^T.
\end{equation}

\subsubsection{Varimax rotation}
After selecting the appropriate number of factors, linear models of normalized primary variables are obtained, which are functions of a certain set of independent (orthogonal) factors. Each factor can be interpreted as a unit orthogonal directional vector of the coordinate system. Then the coefficients of the model of any variable can be treated as a vector in the space of factors. Each standardized primary variable is more or less dependent on all factors.

Since the factors are interpreted as axes of an orthogonal coordinate system, when any orthogonal rotation of this coordinate system is performed, the directions of the axes will change, so the set of factors will change. Consequently, the coordinates of the vectors in the new coordinate system will also change, but the interpretation of the factors will not change. The factors still remain the orthogonal directions of the axes of the coordinate system.

Of the infinitely possible rotations, it would be interesting to see a rotation of the coordinate system after which a given modeled primary variable would be more similar to a single factor and not much similar to the other factors. Similarity to a single factor means that the direction of the vector describing a given primary variable would be as close as possible to the direction of one of the axes of the coordinate system and would be almost orthogonal to the other axes of the coordinate system.

To find such a coordinate system, and thus a set of suitable factors, the Varimax method is used \cite{Kaiser1958}. The Varimax method seeks such a solution that the vector describing the standardized primary variable being modeled has the smallest possible number of factor loadings with large absolute values and a large number of factor loadings with small absolute values. Ideally, the description of a given variable should have one factor loading with an absolute value close to unity, and the remaining factor loadings describing the variable should have absolute values close to zero. After Varimax rotation, each primary variable is usually particularly strongly associated with only one factor or at most a small number of factors.

Formally, the Varimax procedure aims to identify the coordinate system (factors) that maximizes the sum of squares of the dispersion of common variances coming from successive factors. This will increase the number of large and small common variances, while decreasing the number of average common variances.

Common variances are optimized separately on each of the planes defined by a pair of orthogonal axes of the coordinate system. On a given plane formed by two axes of the coordinate system, a rotation of these axes around an axis perpendicular to them is performed. If the rotation operation on a given plane does not increase the value of the objective function, the rotation is skipped, moving to the next plane. Before the rotation procedure, the row vectors of the factor loading matrix are normalized to unit length. After the rotation is completed, the original length of the rotated vectors will be restored \cite{Kaiser1958}.
\subsection{Majority voting -- majority variance rules}
The majority is more than half of the total. A majority of a set, is more than half of the contents of that set. We can say that a majority of a set is a subset consisting of more than half of the elements of the set \cite{Majority2024}.
Absolute majority rule (AMR) is a selection rule that says that when comparing two options, the option preferred by more than half (the majority) of voters wins \cite{Majority_rule2024}. 
An alternative to absolute majority rule is the relative majority rule (RMR). In this case, the winning option is the one that received the most votes, even when a majority of voters would prefer another option \cite{Plurality2024}.

A correct model of a random variable should represent the majority of the variance of the modeled variable. As a criterion for evaluating the quality of a random variable model, both the absolute majority rule and the relative majority rule can be applied to the variance of the variable being modeled. 

\subsubsection{Rule of absolute majority of variance}\label{absolut}
If the model of a variable represents the majority of its variance, then the model represents at least half of its variance.  Analogous to the majority rule used in social elections \cite{Majority2024}\cite{Majority_rule2024}, this can be called the  rule of absolute majority of variance. For example, a majority group of parliamentarians representing more than half of the voters has the ability to make decisions on behalf of all voters. In addition, a model that reproduces more than half of the variance of a modeled variable can be considered a representative model, and the rule of absolute majority of variance can be considered a sufficient condition for a good model. In this article, this rule will be used several times as a binary decision rule.
\subsubsection{Rule of relative majority of variance}\label{relative}
When modeling a random variable, it may be appropriate to consider the relative majority of variance rule instead of the absolute majority of variance rule. The relative majority of variance rule also has its analogy to a decision made in a democratic vote. The relative majority rule can be used when a certain voted proposal receives more votes than any other proposal, but does not receive more than half of all votes \cite{Plurality2024}. 

The rule of relative majority of variance is weaker than the rule of absolute majority of variance. For this reason, it does not guarantee that the resulting random variable model will sufficiently reproduce the variance of the modeled variable. However, its use can be justified in situations where the stronger rule of absolute majority of variance will hide some interesting properties of the modeled variable. Using a weaker rule of relative majority of variance in such a case may uncover these interesting and less obvious properties. 
While the rule of relative majority of variance is not a sufficient condition for a good model, it is a necessary condition for a good model.

\subsection{Similarity of random variables}
The similarity of elements in a set can be discussed in the context of a mathematical similarity relation, which is reflexive and symmetric. The similarity relation divides a set into similarity classes. Belonging to a similarity class means that the elements belonging to the class have at least one characteristic in common, so that they can be said to be indistinguishable \cite{Pogonowski}\cite{Pogonowski1997}.
This situation occurs in factor analysis, where similar variables to factors are identified. The similarity relation is defined at the union of two sets. 
One set is the set of modeled random variables, and the other set is the set of independent factors.

Variables share their variance with factors. Some variables share a significant portion of their variance with some factors. If the absolute (or relative) majority of the variance of a certain variable is modeled by a certain factor, then the variable can be said to be absolutely (or relatively) similar to the factor. If several variables are absolutely (or relatively) similar to the same factor, then these variables belong to the same similarity class.

It should be noted here that absolute similarity of a variable to a factor means that the rule of absolute majority of variance has been used to find similarity. On the other hand, relative similarity of a variable to a factor means that the relative majority of variance rule was used to find similarity.

\section{Algorithm for clustering random attributes}\label{clustAlg}
If the set of attributes to be clustered contains nominal attributes, two problems must be solved for these nominal attributes:
\begin{enumerate}[nosep]
\item The values of a nominal attribute should be encoded using numbers. If there are no classes with identical cardinalities in the set of attribute values, such an attribute can be encoded using the cardinalities of individual classes, according to formula (\ref{ranga}). Without losing any important information contained in the encoded values, the above formula can be simplified.
\item When a nominal attribute contains classes of values with identical cardinality, it can be encoded using complex numbers \cite{Gniazdowski2015}.  However, for nominal attributes encoded with complex numbers, it is not possible to calculate correlations (Subsection \ref{cod2cor}). Therefore, other encoding methods that do not use class cardinality should be used.
\end{enumerate}
After solving both of the above problems, it will be possible to present a universal algorithm for clustering them. This algorithm will effectively cluster both numeric attributes and nominal attributes encoded by numbers.
\subsection{Possible simplification in encoding of nominal data}
Subsection \ref{numCod} adopts a method for encoding nominal data that is analogous to that of rank with tied ranks. A class of identical elements with cardinality $n$ is assigned a rank (\ref{ranga}). The rank assignment algorithm can be written in three steps:
\begin{enumerate}[nosep]
\item Different nominal values are assigned their cardinalities.
\item All cardinalities are divided by two.
\item The number 0.5 is added to the result obtained.
\end{enumerate}
The above algorithm for encoding nominal data is equivalent to the linear transformation of a random attribute, in which nominal values are encoded using class cardinalities.
On the other hand, it was noted that the linear transformation of a random attribute does not change the direction of the vector representing the random component of this attribute, and consequently does not affect the value of the correlation coefficient (Subsections \ref{lin2vect} and \ref{lin2corel}). 
Therefore, for the study of correlation, it is not necessary to encode nominal random attributes with  the formula (\ref{ranga}), and it is sufficient to encode them with the cardinalities of each class:
\begin{equation}
R=n.
\end{equation}
\subsection{Coding of nominal random attributes containing classes with identical cardinalities}
If a nominal attribute contains equicardinal classes of different nominal values, then this attribute can be encoded using complex numbers \cite{Gniazdowski2015}.  
On the other hand, Subsection \ref{cod2cor}   notes that for random attributes that are encoded with complex numbers, it is not possible to compute correlations.
For this reason, in the present work, if any nominal attribute contains equicardinal classes of identical nominal values, the so-called one-hot encoding will be used instead of encoding this attribute with complex numbers \cite{Cerda2018}\cite{Hancock2020}\cite{deepAIglossary}.
The proposed procedure is that each class is encoded as a separate binary random attribute. For each class belonging to a given attribute, a binary column is created with ones in those positions where the value of the attribute belongs to the class. In the remaining positions of this new column, zeros are inserted.
If an attribute takes $k$ different nominal values, it will be replaced by $k$ binary columns, each representing one class. Each class belonging to the attribute is represented as an additional attribute with binary values. As a result, it is now a dichotomous attribute. In general, it should be noted that:
\begin{itemize}[nosep]
\item The rank of a dichotomous random attribute is also a linear transformation. Therefore, both a binary random attribute and a random attribute in which zeros and ones are replaced by ranks are equivalent from the point of view of correlation testing.
\item If the value classes of a dichotomous attribute have different cardinalities, then also replacing these values by their cardinalities is a linear transformation. Thus, such a transformation also does not change (at least the modulus of) the correlation coefficient. 
\end{itemize}

\begin{algorithm}[h!]
\centering
\caption{ Attribute clustering algorithm }\label{algor2}
\fontsize{10}{12}\selectfont{
	\begin{tabular}{r|l }
		Input: &An array with different types of data; \\ 
		\quad \quad  Output:\quad &Graphs of similarity classes between attributes and factors;\quad \quad \quad \quad \quad \quad \quad  \\ \hline 
		Step No. &\quad \quad  Steps of the algorithm\\ \hline 
		1. &Encode nominal data with numbers; \\ 
		2. &Calculate the correlation matrix for numerical data and encoded nominal data; \\ 
		3. &Perform Factor Analysis: \\
		\quad &\quad a.	For the correlation matrix, solve the eigenproblem; \\
		\quad &\quad b.	Calculate the factor loadings matrix; \\
		\quad &\quad c.	Calculate the matrix of common variances; \\
		\quad &\quad d.	Select the number of factors; \\
		\quad &\quad e.	Perform factor rotation; \\ 
		4. \quad &Perform clustering of attributes based on one of the rules: \\
		\quad &\quad a.	The absolute majority of variance rule; \\
		\quad &\quad b.	The relative majority of variance rule. \\
		5. \quad &Draw graphs of similarity classes
	\end{tabular}
}
\end{algorithm}

\subsection{Attribute clustering algorithm}
In the preliminaries, it was shown that nominal data can be encoded using numbers (Subsection \ref{numCod}). If a linear order relation can be defined in the encoded set of numbers , then correlations can also be calculated for the encoded nominal data. If correlations can be calculated, nominal attributes can also be clustered using factor analysis. Factor analysis will divide numerical attributes and encoded nominal attributes into similarity classes.  Each similarity class will contain those random attributes that are similar to the factor. The measure of similarity will be the square of the factor loading, given as a percentage, which measures the level of common variance of the attribute and the factor. The assignment of an attribute to a given similarity class will be determined by the rule of absolute majority of variance (Subsection \ref{absolut}) or the rule of relative majority of variance (Subsection \ref{relative}):
\begin{itemize}[nosep]
\item According to the absolute majority of variance rule, a majority of the common variance is nothing more than more than $50\%$ of the common variance. More than $50\%$ of the common variance means that most of the variance of a given modeled primary attribute is represented by a given factor. In other words, a given modeled attribute is similar to a factor. If more than one modeled attribute is similar to the same factor, then all attributes similar to that factor can be considered to belong to one similarity class and thus form a cluster.
\item According to the rule of relative majority of variance, the assignment to a given cluster will be determined by the level of common variance of the attribute and the factor, related to the total level of variance reproduced by all selected factors. If this level exceeds half of the total reproduced variance of the attribute, then even if it is less than $50\%$, the attribute is considered similar to the factor. For example, if all the selected factors reproduce $70\%$ of the variance of a given attribute, and a certain single factor reproduces $45\%$ of the variance of that attribute, then the attribute is considered similar to the factor, despite the fact that the factor does not represent the majority of its variance. In the example given here, selected factors are considered to be responsible for $70\%$ of the variance of a given attribute, and rejected factors, which can be treated as noise, are responsible for $30\%$ of this variance.  Of this $70\%$ of the variance, as much as $45\%$ of the variance is accounted for by a single factor. It can be seen that its influence on a given attribute dominates the influence of the other factors.  Therefore, an attribute can be considered similar to it.
\end{itemize}
Clustering of random attributes is described in the form of Algorithm \ref{algor2}. 
The algorithm presented here is universal because it works on different types of data: on numerical data, on ordinal data represented by numbers, and on nominal data encoded with numbers. In the following sections of this article, the operation of this algorithm will be shown through several examples, for different data sets taken from different sources. First, examples operating only on nominal data will be shown. Later, examples will also be shown with data that contain random attributes measured at different measurement scales: numeric, ordinal, nominal.
\section{Examples of the application of the attribute clustering algorithm}\label{examples}
The algorithm for clustering numerical attributes and/or numerically encoded nominal attributes will be demonstrated with examples.
The first example will bebased on a small data set. 
This will allow to follow in detail the operation of the proposed algorithm as well as the presentation of its results. 
The remaining examples will focus on presenting the results of the algorithm.

Regardless of the scales on which the input data were measured, after nominal data are encoded with numbers, all input data are represented in numerical form.
Data measured on ratio, interval, and ordinal scales can be ordered linearly.
On the other hand, both cardinality-encoded data and one-hot-encoded data can also be linearly ordered. Therefore, all such data can also be ranked.
In the following examples, the effect of ranking on the clustering results will be tested. For this purpose, two more cases of clustering will be considered:
\begin{itemize}[nosep]
\item In the first case without ranking, both numeric attributes and encoded nominal attributes will be clustered.
\item In the second case, all attributes measured on numerical scales ( ratio, interval, ordinal), as well as non-dichotomous nominal attributes encoded by class cardinalities, will be ranked before the clustering procedure.  Since for a dichotomous set, its ranking is equivalent to some linear transformation that does not affect the value of the correlation coefficient, therefore dichotomous attributes, including attributes representing classes, obtained by the one-hot encoding method will not be ranked.
\end{itemize}
By analyzing the clustering results obtained for unranked attributes, as well as the clustering results for ranked attributes, the effect of rank on the clustering result will be studied. 
However, before applying the algorithm to cluster attributes in different datasets, it is necessary to introduce the attribute name encoding used in the algorithm. On the one hand, attribute names consist of several letters of the alphabet and can therefore be relatively long. On the other hand, one-hot encoding creates new attributes whose names should contain information about the encoded value class as well as information about the name of the attribute from which the class is derived. The use of a name encoding system will facilitate the interpretation of both the intermediate and final results of the algorithm. However, for this to happen, the method used to encode the names must first be explained.

\subsection{Encoding of attribute names}\label{codName}
To avoid excessively long strings in attribute names or one-hot encoded class names, an abbreviated name system is used to encode them. 
Successive columns of the table with encoded data (successive random attributes) will be denoted by symbols of the form $Ak$ or \textit{Ak>m},
where $A$ is the first letter of the alphabet, and $k$ and $m$ are positive integers.
The number $k$ occurring after the letter $A$ is the number of the random attribute from the set with the original data. 
The number $k$ takes values from $1$ to $n$, where $n$ is the number of all random attributes in the original dataset.

The attribute name in the form of $Ak$ (without the suffix ''\textit{>m}'') indicates that the column represents a random numeric attribute or a random nominal attribute that does not have equicardinal classes of identical values. Such a nominal attribute has been encoded with class cardinalities. 
If the symbol ''$Ak$'' is followed by the suffix ''\textit{>m}'' in the attribute name, it means that:
\begin{itemize}[nosep]
\item The $k$-th attribute contains equicardinal classes with identical nominal values. 
\item The one-hot encoding method was used to encode the $k$-th attribute.
\end{itemize}
The suffix ''\textit{>m}'' occurring after the symbol ''$Ak$'' indicates the number of the $m$-th class of  nominal values contained in the $k$-th attribute. This class is represented as a separate binary column.

On the other hand, in order to simplify the interpretation of clustering (e.g., the results of clustering shown in the chart), it will be convenient to use the full names found in the original data before encoding. In this case, the attribute names after encoding will have one of the following two forms: \sloppy
''\textit{Attribute\_Name}'' or ''\textit{Attribute\_Name>Class\_Name}''. The first form will refer only to attributes that do not contain equicardinal classes with identical values. 
The second form will refer to attributes that contain equicardinal classes with identical nominal values.

\begin{table}[h!]
\centering
\caption{\textit{Simple Weather Forecast} dataset \cite{weather2022}}\label{weather}
\fontsize{8.5}{12}\selectfont{
	\begin{tabular}{|c|c|c|c|c|}
		\hline
		outlook&temperature&humidity&windy&play\\ \hline
		sunny&hot&high&false&no\\
		sunny&hot&high&true&no\\
		overcast&hot&high&false&yes\\
		rainy&mild&high&false&yes\\
		rainy&cool&normal&false&yes\\
		rainy&cool&normal&true&no\\
		overcast&cool&normal&true&yes\\
		sunny&mild&high&false&no\\
		sunny&cool&normal&false&yes\\
		rainy&mild&normal&false&yes\\
		sunny&mild&normal&true&yes\\
		overcast&mild&high&true&yes\\
		overcast&hot&normal&false&yes\\
		rainy&mild&high&true&no\\
		\hline
	\end{tabular}
}
\end{table}

\begin{table}[h!]
\centering
\caption{\textit{Simple Weather Forecast} dataset -- naming schemes applied to attributes in Table \ref{weather}}\label{weatherCodComp}
\fontsize{8.5}{12}\selectfont{
	\begin{tabular}{|c|c|c|} 	\hline
		No.&Full name of the encoded attribute&Short name of the encoded attribute \\ \hline
		1.&outlook>sunny &A1>1 \\
		2.&outlook>overcast &A1>2 \\
		3.&outlook>rainy &A1>3 \\
		4.&temperature>hot &A2>1 \\
		5.&temperature>mild &A2>2 \\
		6.&temperature>cool &A2>3 \\
		7.&humidity>high &A3>1 \\
		8.&humidity>normal &A3>2 \\
		9.&windy &A4 \\
		10.&play&A5
		\\
		\hline
	\end{tabular}
}
\end{table}

\begin{table}[h!]
\centering
\caption{Coded dataset \textit{Simple Weather Forecast}}\label{weatherCod}
\fontsize{8.5}{12}\selectfont{
	\begin{tabular}{|c|c|c|c|c|c|c|c|c|c|} 	\hline
		\multicolumn{3}{|c|}{outlook} & \multicolumn{3}{|c}{temperature} & \multicolumn{2}{|c|}{humidity}&windy&play\\ \hline
		overcast&rainy&sunny&hot&mild&cool&high&normal&false/true&yes/no\\ \hline
		\textit{A1>1}&\textit{A1>2}&\textit{A1>3}&\textit{A2>1}&\textit{A2>2}&\textit{A2>3}&\textit{A3>1}&\textit{A3>2}&\textit{A4}&\textit{A5}\\ \hline
		1&0&0&1&0&0&1&0&8&5\\
		1&0&0&1&0&0&1&0&6&5\\
		0&1&0&1&0&0&1&0&8&9\\
		0&0&1&0&1&0&1&0&8&9\\
		0&0&1&0&0&1&0&1&8&9\\
		0&0&1&0&0&1&0&1&6&5\\
		0&1&0&0&0&1&0&1&6&9\\
		1&0&0&0&1&0&1&0&8&5\\
		1&0&0&0&0&1&0&1&8&9\\
		0&0&1&0&1&0&0&1&8&9\\
		1&0&0&0&1&0&0&1&6&9\\
		0&1&0&0&1&0&1&0&6&9\\
		0&1&0&1&0&0&0&1&8&9\\
		0&0&1&0&1&0&1&0&6&5
		\\
		\hline
	\end{tabular}
}
\end{table}

\subsection{Dataset No. 1 -- \textit{Simple Weather Forecast}}\label{Dat1}
The nominal attribute clustering algorithm will be presented in detail with an example. 
The algorithm will be tested using the \textit{Simple Weather Forecast} dataset, which is available on the Kaggle platform \cite{weather2022}.

It can be noted that in the \textit{Simple Weather Forecast} dataset, the values of some attributes can be ordered. Therefore, they can be assigned numerical values. This group of attributes includes the temperature attribute ($hot>mild>cool$) and the humidity attribute ($high>normal$). The possibility of ordering the forecast attribute ($sunny>weather>rainy$) is also not excluded. 
However, the author of the article abandons the possibility of ordering the values of these attributes, treating them consistently as nominal attributes. As a result, the example discussed here becomes a dummy that plays a didactic role.
The approach adopted here will make it possible to show how nominal data can be encoded with numbers, both by the one-hot method and by class cardinality.
So, the \textit{Simple Weather Forecast} dataset contains nominal data describing $14$ points in the space of five random attributes (Table \ref{weather}):
\begin{itemize}[nosep]
\item The first attribute (\textit{outlook}) takes three different random values: \textit{sunny}, \textit{overcast}, \textit{rainy}. \textit{Sunny} and \textit{rainy} values appear 5 times in the input dataset. Thus, the \textit{outlook} attribute contains two equicardinal classes. 
\item The second attribute (\textit{temperature}) also takes three different values: \textit{hot}, \textit{mild}, \textit{cool}. The values \textit{hot} and \textit{cool} appear four times. So, these values form two equicardinal classes.
\item The third attribute (\textit{humidity}) takes two values: \textit{high} and \textit{normal}. Both of these values appear 7 times in the input data set, thus forming 2 equicardinal classes.
\item The fourth attribute (\textit{windy}) takes two values: \textit{false} and \textit{true}. In this case, no equicardinal classes appear.
\item The fifth and final attribute (\textit{play}) takes two values: \textit{no} and \textit{yes}. Here, too, there are no equicardinal classes.				
\end{itemize}
The method of encoding attribute and class names presented in subsection \ref{codName} was applied to the dataset presented in Table \ref{weather}.
Table \ref{weatherCodComp} shows both of the above name encoding schemes applied to the data in Table \ref{weather}. 
Due to the existence of classes with equal cardinalities, the first three attributes were encoded using the one-hot encoding method. The last two attributes do not have value classes with equal cardinalities, so they were encoded with class cardinalities. 
Although the original data was in five columns, due to the need to use the one-hot encoding method, after encoding the data was stored in ten columns (Table \ref{weatherCod}).

\begin{table}[h!]
\centering
\caption{\textit{Simple Weather Forecast} dataset -- correlation matrix for encoded random attributes}\label{weatherCorel}
\fontsize{8.5}{12}\selectfont{
	\begin{tabular}{|c|c|c|c|c|c|c|c|c|c|c|} 	\hline
		&\textit{A1>1}&\textit{A1>2}&\textit{A1>3}&\textit{A2>1}&\textit{A2>2}&\textit{A2>3}&\textit{A3>1}&\textit{A3>2}&\textit{A4}&\textit{A5}\\ \hline
		\textit{A1>1}&1.000&-0.471&-0.556&0.189&-0.043&-0.141&0.149&-0.149&0.043&-0.378\\
		\textit{A1>2}&-0.471&1.000&-0.471&0.300&-0.228&-0.050&0.000&0.000&-0.091&0.471\\
		\textit{A1>3}&-0.556&-0.471&1.000&-0.471&0.258&0.189&-0.149&0.149&0.043&-0.067\\
		\textit{A2>1}&0.189&0.300&-0.471&1.000&-0.548&-0.400&0.316&-0.316&0.228&-0.189\\
		\textit{A2>2}&-0.043&-0.228&0.258&-0.548&1.000&-0.548&0.289&-0.289&-0.125&0.043\\
		\textit{A2>3}&-0.141&-0.050&0.189&-0.400&-0.548&1.000&-0.632&0.632&-0.091&0.141\\
		\textit{A3>1}&0.149&0.000&-0.149&0.316&0.289&-0.632&1.000&-1.000&0.000&-0.447\\
		\textit{A3>2}&-0.149&0.000&0.149&-0.316&-0.289&0.632&-1.000&1.000&0.000&0.447\\
		\textit{A4}&0.043&-0.091&0.043&0.228&-0.125&-0.091&0.000&0.000&1.000&0.258\\
		\textit{A5}&-0.378&0.471&-0.067&-0.189&0.043&0.141&-0.447&0.447&0.258&1.000
		\\
		\hline
	\end{tabular}
}
\end{table}	

\begin{table}[h!]
\centering
\caption{\textit{Simple Weather Forecast} dataset -- matrix of coefficients of determination for encoded random attributes}\label{weatherDeterm}
\fontsize{8.5}{12}\selectfont{
	\begin{tabular}{|c|c|c|c|c|c|c|c|c|c|c|} 	\hline
		&\textit{A1>1}&\textit{A1>2}&\textit{A1>3}&\textit{A2>1}&\textit{A2>2}&\textit{A2>3}&\textit{A3>1}&\textit{A3>2}&\textit{A4}&\textit{A5}\\ \hline
		\textit{A1>1}&100\%&22.2\%&30.9\%&3.6\%&0.2\%&2.0\%&2.2\%&2.2\%&0.2\%&14.3\%\\
		\textit{A1>2}&22.2\%&100\%&22.2\%&9.0\%&5.2\%&0.3\%&0.0\%&0.0\%&0.8\%&22.2\%\\
		\textit{A1>3}&30.9\%&22.2\%&100\%&22.2\%&6.7\%&3.6\%&2.2\%&2.2\%&0.2\%&0.4\%\\
		\textit{A2>1}&3.6\%&9.0\%&22.2\%&100\%&30.0\%&16.0\%&10.0\%&10.0\%&5.2\%&3.6\%\\
		\textit{A2>2}&0.2\%&5.2\%&6.7\%&30.0\%&100\%&30.0\%&8.3\%&8.3\%&1.6\%&0.2\%\\
		\textit{A2>3}&2.0\%&0.3\%&3.6\%&16.0\%&30.0\%&100\%&40.0\%&40.0\%&0.8\%&2.0\%\\
		\textit{A3>1}&2.2\%&0.0\%&2.2\%&10.0\%&8.3\%&40.0\%&100\%&100\%&0.0\%&20.0\%\\
		\textit{A3>2}&2.2\%&0.0\%&2.2\%&10.0\%&8.3\%&40.0\%&100\%&100\%&0.0\%&20.0\%\\
		\textit{A4}&0.2\%&0.8\%&0.2\%&5.2\%&1.6\%&0.8\%&0.0\%&0.0\%&100\%&6.7\%\\
		\textit{A5}&14.3\%&22.2\%&0.4\%&3.6\%&0.2\%&2.0\%&20.0\%&20.0\%&6.7\%&100.0\%
		\\
		\hline
	\end{tabular}
}
\end{table}

\begin{table}[h!]
\centering
\caption{\textit{Simple Weather Forecast} dataset -- eigenvalues of the matrix of correlation coefficients}\label{weatherEigval}
\fontsize{8.5}{12}\selectfont{
	\begin{tabular}{|c|c|c|c|c|c|c|c|c|c|c|} 	\hline
		Eigenvalue no.&1&2&3&4&5&6&7&8&9&10\\ 	\hline
		Eigenvalue&3.18&2.18&1.70&1.16&1.09&0.50&0.19&3.3E-16&2.2E-16&0
		\\
		\hline
	\end{tabular}
}
\end{table}

\begin{table}[h!]
\centering
\caption{\textit{Simple Weather Forecast} dataset -- eigenvectors in columns}\label{weatherEigvect}
\fontsize{8.5}{12}\selectfont{
	\begin{tabular}{|c|c|c|c|c|c|c|c|c|c|} 	\hline
		\textit{V1}&\textit{V2}&\textit{V3}&\textit{V4}&\textit{V5}&\textit{V6}&\textit{V7}&\textit{V8}&\textit{V9}&\textit{V10}\\ 	\hline
		-0.230&0.072&0.564&0.060&-0.496&0.082&0.144&-7.8E-05&0.588&0\\
		0.042&0.448&-0.502&-0.281&0.008&0.117&-0.380&-7.4E-05&0.555&0\\
		0.190&-0.495&-0.091&0.205&0.489&-0.193&0.214&-7.8E-05&0.588&0\\
		-0.275&0.491&0.063&0.150&0.272&-0.473&0.216&0.559&7.4E-05&0\\
		-0.148&-0.493&-0.317&0.011&-0.445&-0.100&-0.228&0.612&8.1E-05&0\\
		0.437&0.049&0.284&-0.162&0.215&0.583&0.034&0.559&7.4E-05&0\\
		-0.514&-0.076&-0.144&-0.034&0.177&0.377&0.187&1.8E-10&3.3E-10&0.707\\
		0.514&0.076&0.144&0.034&-0.177&-0.377&-0.187&-1.8E-10&-3.3E-10&0.707\\
		-0.005&0.133&-0.028&0.878&0.030&0.261&-0.376&1.3E-10&2.4E-10&0\\
		0.303&0.180&-0.440&0.230&-0.366&0.131&0.691&4.1E-11&1.2E-10&0
		\\
		\hline
	\end{tabular}
}
\end{table}

\begin{table}[h!]
\centering
\caption{\textit{Simple Weather Forecast} dataset -- full matrix of factor loadings}\label{weatherFullLoad}
\fontsize{8.5}{12}\selectfont{
	\begin{tabular}{|c|c|c|c|c|c|c|c|c|c|c|} 	\hline
		&$F1$&$F2$&$F3$&$F4$&$F5$&$F6$&$F7$&$F8$&$F9$&$F10$\\ \hline
		\textit{A1>1}&-0.410&0.107&0.736&0.064&-0.517&0.058&0.063&-1.4E-12&8.8E-09&0\\
		\textit{A1>2}&0.075&0.662&-0.655&-0.303&0.008&0.083&-0.167&-1.3E-12&8.3E-09&0\\
		\textit{A1>3}&0.339&-0.731&-0.118&0.221&0.510&-0.136&0.094&-1.4E-12&8.8E-09&0\\
		\textit{A2>1}&-0.490&0.725&0.082&0.161&0.284&-0.335&0.095&1.0E-08&1.1E-12&0\\
		\textit{A2>2}&-0.264&-0.728&-0.413&0.012&-0.463&-0.071&-0.100&1.1E-08&1.2E-12&0\\
		\textit{A2>3}&0.779&0.072&0.370&-0.175&0.224&0.412&0.015&1.0E-08&1.1E-12&0\\
		\textit{A3>1}&-0.916&-0.113&-0.187&-0.036&0.184&0.266&0.082&3.2E-18&4.9E-18&0\\
		\textit{A3>2}&0.916&0.113&0.187&0.036&-0.184&-0.266&-0.082&-3.2E-18&-4.9E-18&0\\
		\textit{A4}&-0.009&0.196&-0.036&0.948&0.032&0.185&-0.165&2.3E-18&3.7E-18&0\\
		\textit{A5}&0.540&0.266&-0.574&0.248&-0.381&0.092&0.304&7.4E-19&1.7E-18&0
		\\
		\hline
	\end{tabular}
}
\end{table}

\begin{table}[h!]
\centering
\caption{\textit{Simple Weather Forecast} dataset -- full matrix of common variances}\label{weatherFullCommon}
\fontsize{8.5}{12}\selectfont{
	\begin{tabular}{|c|c|c|c|c|c|c|c|c|c|c|} 	\hline
		&$F1$&$F2$&$F3$&$F4$&$F5$&$F6$&$F7$&$F8$&$F9$&$F10$\\ \hline
		\textit{A1>1}&0.168&0.011&0.541&0.004&0.267&0.003&0.004&2.0E-24&7.8E-17&0\\
		\textit{A1>2}&0.006&0.439&0.429&0.092&0.000&0.007&0.028&1.8E-24&6.9E-17&0\\
		\textit{A1>3}&0.115&0.535&0.014&0.049&0.260&0.019&0.009&2.0E-24&7.8E-17&0\\
		\textit{A2>1}&0.240&0.526&0.007&0.026&0.081&0.112&0.009&1.0E-16&1.2E-24&0\\
		\textit{A2>2}&0.070&0.530&0.170&0.000&0.215&0.005&0.010&1.2E-16&1.5E-24&0\\
		\textit{A2>3}&0.607&0.005&0.137&0.031&0.050&0.170&0.000&1.0E-16&1.2E-24&0\\
		\textit{A3>1}&0.839&0.013&0.035&0.001&0.034&0.071&0.007&1.0E-35&2.4E-35&0\\
		\textit{A3>2}&0.839&0.013&0.035&0.001&0.034&0.071&0.007&1.0E-35&2.4E-35&0\\
		\textit{A4}&0.000&0.038&0.001&0.898&0.001&0.034&0.027&5.3E-36&1.3E-35&0\\
		\textit{A5}&0.292&0.071&0.329&0.062&0.145&0.009&0.092&5.5E-37&3.1E-36&0
		\\
		\hline
	\end{tabular}
}
\end{table}

\subsubsection{Initial steps before clustering encoded nominal attributes}
For the encoded data in Table \ref{weatherCod}, the calculations necessary to perform clustering of attributes due to their similarity to factors were performed. 
This section will present only those results needed to show how the nominal attribute clustering algorithm presented in Section \ref{clustAlg} works.
First, a matrix of correlation coefficients was calculated for the encoded attributes. 
The results are shown in Table \ref{weatherCorel}.
The mutual similarity of the encoded attributes was then estimated. For this purpose, a matrix of coefficients of determination was calculated. This matrix is shown in Table \ref{weatherDeterm}. 
In the next step, an eigenproblem was solved for the correlation matrix. The calculated eigenvalues are shown in Table \ref{weatherEigval}, while the eigenvectors are shown in Table \ref{weatherEigvect}.

Using the square roots of the eigenvalues, as well as the eigenvector matrix (Table \ref{weatherEigvect}), using the algorithm described in subsection \ref{FA}, the full factor loading matrix was calculated (Table \ref{weatherFullLoad}).
Then, as the Hadamard square of the full factor loading matrix, the common variances between the encoded attributes and the factors were calculated (Table \ref{weatherFullCommon}).

Due to the nature of the data, attribute clustering was implemented only for the encoded data, without ranking. Encoding the first three attributes using the one-hot method created additional binary attributes. The possible ranking of the binary attributes is a linear transformation that does not affect the direction of the random vector. On the other hand, the last two attributes were encoded using class cardinality. Since both attributes contain two classes of values each, they will be encoded as dichotomous sets of numbers. The ranking of a dichotomous set is also a linear transformation that does not affect the direction of the random vector. As a result, ranking the encoded attributes would not change their correlations with each other, and thus could not affect the results of attribute clustering.

\begin{figure}[h!]
\centering
\includegraphics[width=0.75\textwidth]{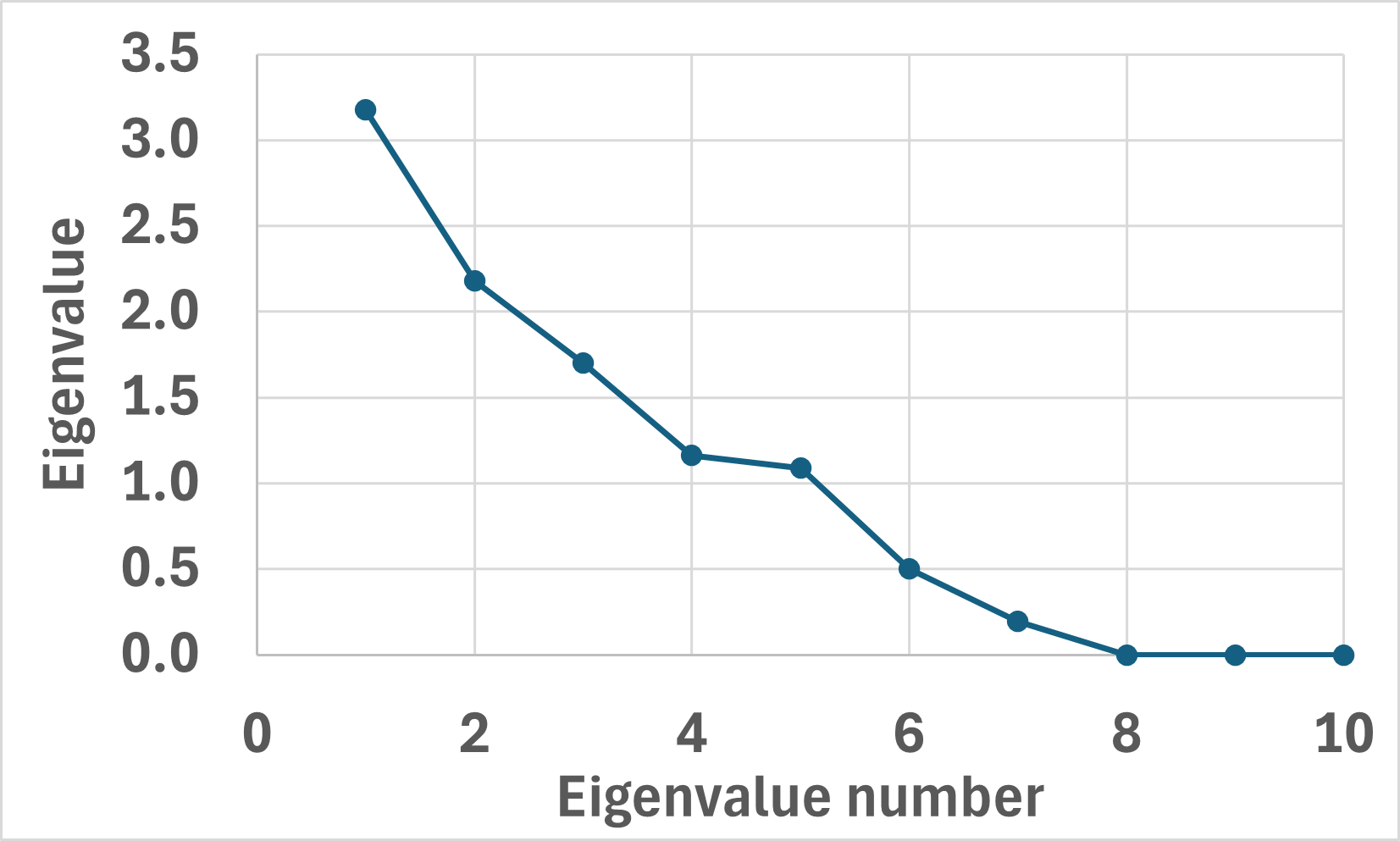}
\caption{Scree plot for \textit{Simple Weather Forecast} dataset}\label{weatherScree}
\end{figure}

\begin{table}[h!]
\centering
\caption{\textit{Simple Weather Forecast} dataset -- percentage of the variance explained by successive factors}\label{weatherAverCriterium}
\fontsize{8.5}{12}\selectfont{
	\begin{tabular}{c|c|c|c|c} \hline 
		Factor & \multirow{2}{*}{Eigenvalue} & Cumulative & Percentage of the variance & Cumulative percentage  \\ 
		no. & & eigenvalues & explained by each factor &  of explained variance \\ \hline 
		1&3.18&3.18&31.8\%&31.8\%\\
		2&2.18&5.36&21.8\%&53.6\%\\
		3&1.70&7.06&17.0\%&70.6\%\\
		4&1.16&8.22&11.6\%&82.2\%\\
		5&1.09&9.31&10.9\%&93.1\%\\
		6&0.50&9.81&5.0\%&98.1\%\\
		7&0.19&10&1.9\%&100\%\\
		8&0&10&0.0\%&100\%\\
		9&0&10&0.0\%&100\%\\
		10&0&10&0.0\%&100\%
		\\ \hline
	\end{tabular}
}
\end{table}

\begin{table}[h!]
\centering
\caption{\textit{Simple Weather Forecast} dataset -- cumulative matrix of common variances}\label{weatherCumulCommon}
\fontsize{8.5}{12}\selectfont{
	\begin{tabular}{|c|c|c|c|c|c|c|c|c|c|c|} 	\hline
		&$F1$&$F2$&$F3$&$F4$&$F5$&$F6$&$F7$&$F8$&$F9$&$F10$\\ \hline
		\textit{A1>1}&0.168&0.180&0.721&0.725&0.993&0.996&1&1&1&1\\
		\textit{A1>2}&0.006&0.444&0.873&0.965&0.965&0.972&1&1&1&1\\
		\textit{A1>3}&0.115&0.650&0.664&0.713&0.973&0.991&1&1&1&1\\
		\textit{A2>1}&0.240&0.766&0.772&0.799&0.879&0.991&1&1&1&1\\
		\textit{A2>2}&0.070&0.600&0.770&0.770&0.985&0.990&1&1&1&1\\
		\textit{A2>3}&0.607&0.612&0.749&0.780&0.830&1.000&1&1&1&1\\
		\textit{A3>1}&0.839&0.852&0.887&0.888&0.922&0.993&1&1&1&1\\
		\textit{A3>2}&0.839&0.852&0.887&0.888&0.922&0.993&1&1&1&1\\
		\textit{A4}&0.000&0.038&0.040&0.938&0.939&0.973&1&1&1&1\\
		\textit{A5}&0.292&0.363&0.692&0.754&0.899&0.908&1&1&1&1\\ \hline
		Average &\multirow{2}{*}{ 0.318}&\multirow{2}{*}{ 0.536}&\multirow{2}{*}{ 0.706}&\multirow{2}{*}{ 0.822}&\multirow{2}{*}{ 0.931}&\multirow{2}{*}{ 0.981}&\multirow{2}{*}{ 1}&\multirow{2}{*}{ 1}&\multirow{2}{*}{ 1}&\multirow{2}{*}{ 1}\\
		in column&&&&&&&&&&\\ \hline
		Minimum &\multirow{2}{*}{ 0.000}&\multirow{2}{*}{ 0.038}&\multirow{2}{*}{ 0.040}&\multirow{2}{*}{ 0.713}&\multirow{2}{*}{ 0.830}&\multirow{2}{*}{ 0.908}&\multirow{2}{*}{ 1}&\multirow{2}{*}{ 1}&\multirow{2}{*}{ 1}&\multirow{2}{*}{ 1}\\
		in column&&&&&&&&&&
		\\
		\hline
	\end{tabular}
}
\end{table}

\begin{table}[h!]
\centering
\caption{\textit{Simple Weather Forecast} dataset -- given as a percentage: successive eigenvalues (ScreePlt) normalized to the trace of the correlation matrix, as well as two variances represented by successive factors: minimum variance (MinVar) and average variance (AverVar)}\label{weatherMinVar}
\fontsize{8.5}{12}\selectfont{
	\begin{tabular}{|c|c|c|c|c|c|c|c|c|c|c|} 	\hline
		Factors&1&2&3&4&5&6&7&8&9&10\\ \hline
		ScreePlt&31.8\%&21.8\%&17.0\%&11.6\%&10.9\%&5.0\%&1.9\%&0\%&0\%&0\% \\ 
		MinVar&0.0\%&3.8\%&4.0\%&71.3\%&83.0\%&90.8\%&100\%&100\%&100\%&100\%\\ 
		AverVar&31.8\%&53.6\%&70.6\%&82.2\%&93.1\%&98.1\%&100\%&100\%&100\%&100\%\\ 
		MinVarId&\textit{A4}&\textit{A4}&\textit{A4}&\textit{A1>3}&\textit{A2>3}&\textit{A5}&\textit{A5}&\textit{A5}&\textit{A5}&\textit{A5}
		\\
		\hline
	\end{tabular}
}
\end{table}

\begin{figure}[h!]
\centering
\includegraphics[width=0.75\textwidth]{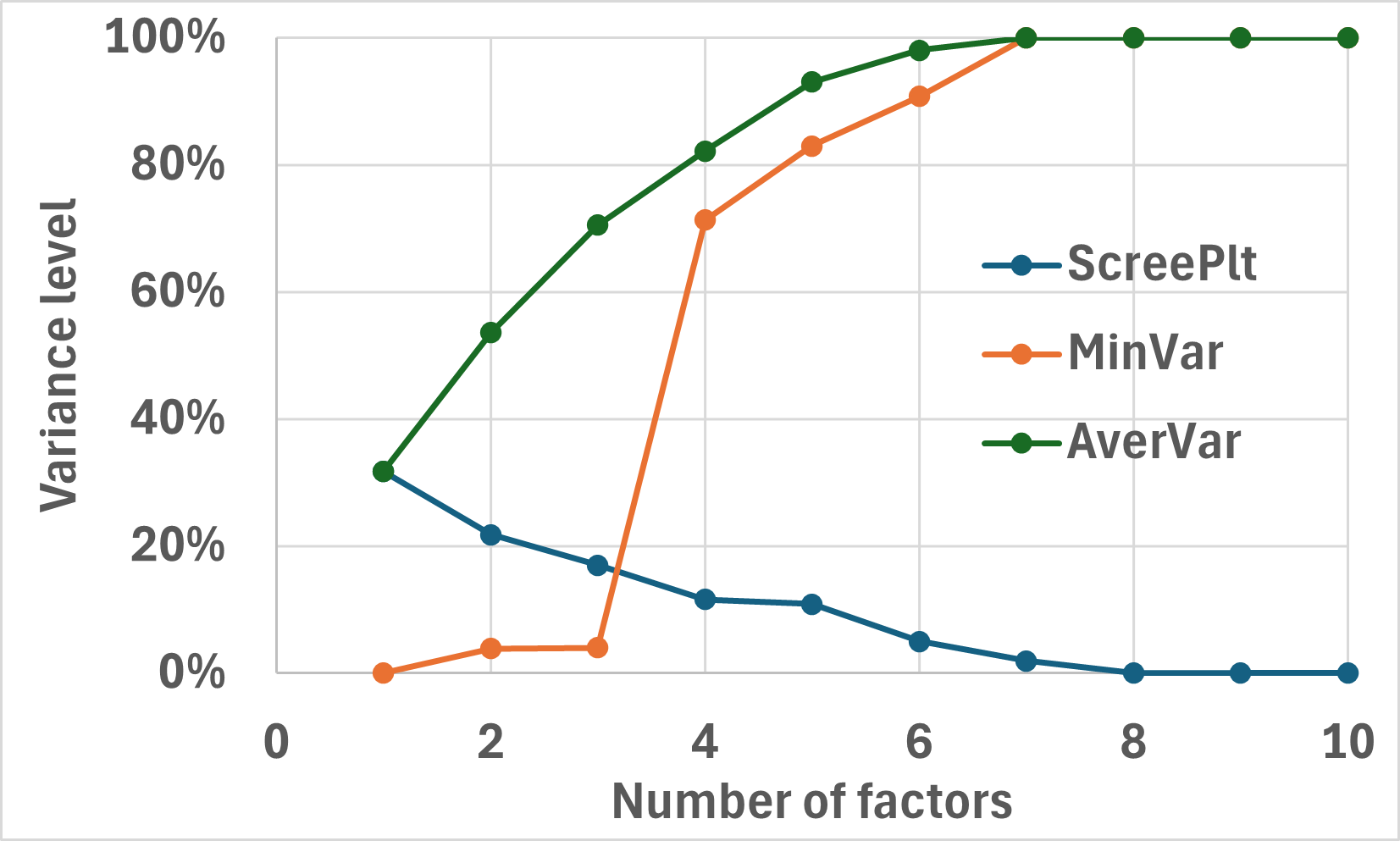}
\caption{\textit{Simple Weather Forecast} dataset -- minimum variance (MinVar) and average variance (AverVar) of attributes, reconstructed by successive factors, shown against a normalized scree plot}\label{weatherMinVarPlot}
\end{figure}

\begin{table}[h!]
\centering
\caption{\textit{Simple Weather Forecast} dataset -- final factor loading matrix}\label{weatherFinFact}
\fontsize{8.5}{12}\selectfont{
	\begin{tabular}{|c|c|c|c|c|c|c|} 	\hline
		&$F1$&$F2$&$F3$&$F4$&Common variances\\ \hline
		\textit{A1>1}&-0.410&0.107&0.736&0.064&72.5\%\\
		\textit{A1>2}&0.075&0.662&-0.655&-0.303&96.5\%\\
		\textit{A1>3}&0.339&-0.731&-0.118&0.221&71.3\%\\
		\textit{A2>1}&-0.490&0.725&0.082&0.161&79.9\%\\
		\textit{A2>2}&-0.264&-0.728&-0.413&0.012&77.0\%\\
		\textit{A2>3}&0.779&0.072&0.370&-0.175&78.0\%\\
		\textit{A3>1}&-0.916&-0.113&-0.187&-0.036&88.8\%\\
		\textit{A3>2}&0.916&0.113&0.187&0.036&88.8\%\\
		\textit{A4}&-0.009&0.196&-0.036&0.948&93.8\%\\
		\textit{A5}&0.540&0.266&-0.574&0.248&75.4\%
		\\ \hline
\end{tabular}}
\end{table}

\begin{table}[h!]
\centering
\caption{\textit{Simple Weather Forecast} dataset -- final matrix of common variances}\label{weatherFinCommon}
\fontsize{8.5}{12}\selectfont{
	\begin{tabular}{|c|c|c|c|c|c|c|} 	\hline
		&$F1$&$F2$&$F3$&$F4$&Common variances\\ \hline
		\textit{A1>1}&16.8\%&1.1\%&\textbf{54.1\%}&0.4\%&72.5\%\\
		\textit{A1>2}&0.6\%&43.9\%&42.9\%&9.2\%&96.5\%\\
		\textit{A1>3}&11.5\%&\textbf{53.5\%}&1.4\%&4.9\%&71.3\%\\
		\textit{A2>1}&24.0\%&\textbf{52.6\%}&0.7\%&2.6\%&79.9\%\\
		\textit{A2>2}&7.0\%&\textbf{53.0\%}&17.0\%&0.0\%&77.0\%\\
		\textit{A2>3}&\textbf{60.7\%}&0.5\%&13.7\%&3.1\%&78.0\%\\
		\textit{A3>1}&\textbf{83.9\%}&1.3\%&3.5\%&0.1\%&88.8\%\\
		\textit{A3>2}&\textbf{83.9\%}&1.3\%&3.5\%&0.1\%&88.8\%\\
		\textit{A4}&0.0\%&3.8\%&0.1\%&\textbf{89.8\%}&93.8\%\\
		\textit{A5}&29.2\%&7.1\%&32.9\%&6.2\%&75.4\%
		\\ \hline
	\end{tabular}
}
\end{table}

\subsubsection{Selection of factors}
The next step proceeded to select the appropriate number of factors. For this purpose, it was checked whether the scree plot (Fig. \ref{weatherScree}) would allow to unambiguously determine the number of factors. 
In the scree plot shown, a clear elbow dividing the graph into two distinct phases cannot be seen. 
In view of the ambiguity of the scree plot criterion, the percentage criterion of the part of the variance explained by the factors was used to find the right number of factors. 
It was assumed that the factor model should reproduce at least $70\%$ of the variance of the modeled attributes. It was read from Table \ref{weatherAverCriterium} that three factors were sufficient to represent at least $70\%$ of the variance of the modeled attributes, explaining $70.6\%$ of the modeled variance.

The applied percentage criterion of the variance explained by the factors provides a guarantee that the three factors will reproduce more than $70\%$ of the average variance of the modeled attributes. However, the same criterion does not guarantee that the three factors will reproduce most of the variance of each individual  attribute modeled. To test this, the Algorithm  \ref{algor1} was used. First, the cumulative common variance matrix was calculated (Table \ref{weatherCumulCommon}).  The bottom two rows of this matrix contain the calculated average values in each column and the found minimum values in each column. Note that the calculated average values are identical to the percentages given in the fourth column in Table \ref{weatherAverCriterium}, which contains the cumulative percentage of explained variance.

As a result, it can be seen that the three factors selected on the basis of the percentage criterion of common variance, which explain more than $70\%$ of the average variance of the modeled attributes, in the case of the  attribute with the designation \textit{A4} explain only about $4\%$ of the variance of this  attribute. It follows that the percentage criterion of common variance is not sufficient to select the right number of factors. The selection of the number of factors should take into account the ability to reproduce most of the variance of all individual  attributes, not just their average variance.
To properly solve the problem of selecting the number of factors, a Table \ref{weatherMinVar} was created based on the cumulative common variance matrix (Table \ref{weatherCumulCommon}).
This table contains the following rows:
\begin{itemize}[nosep]
\item Factors -- the number of factors considered in the factor model,
\item ScreePlt -- successive eigenvalue normalized to the trace of the correlation matrix (Table \ref{weatherCorel}), identical to the corresponding value in the fourth (penultimate) column in Table \ref{weatherAverCriterium},
\item MinVar -- the minimum variance of some attribute, reproduced by successive factors,
\item AverVar -- the cumulative average variance of the modeled attributes, identical to the corresponding value in the last column in Table \ref{weatherAverCriterium},
\item MinVarId -- the identifier of the attribute that is least represented by the factor model. The variance of this attribute reconstructed by successive factors is equal to the MinVar value.
\end{itemize}
Based on the contents of Table \ref{weatherMinVar}, the graph shown in Figure \ref{weatherMinVarPlot} was created, which shows:
\begin{itemize}[nosep]
\item The successive eigenvalues, normalized to the trace of the correlation matrix (ScreePlt), which form a graph that is identical in shape to the scree plot shown in Figure \ref{weatherScree}).
\item The value of the cumulative average variance (AverVar), which increases with successive factors to unity, 
\item The value of the minimum variance (Minvar) that is reproduced by the successive factors.
\end{itemize}
Based on Table \ref{weatherMinVar} and Figure \ref{weatherMinVarPlot}, it can be concluded that only four factors are sufficient to reproduce most (i.e., no less than the assumed 55\%) of the variance of all modeled attributes. 
Evidence of this can be read in the column of Table \ref{weatherMinVar}, which refers to four factors. Thus, the four factors reproduce $82.2\%$ of the average variance of all attributes (AverVar).
On the other hand, the least represented attribute, which was assigned the identifier \textit{A1>3} (MinVarId), has a reconstructed variance level of $71.3\%$ (MinVar).

After selecting the appropriate number of factors, the factor loadings matrix was reduced, as well as the common variance matrix. For the four selected factors, the factor loadings matrix is shown in Table \ref{weatherFinFact}, and the corresponding common variance matrix is shown in Table \ref{weatherFinCommon}.

In Table \ref{weatherFinCommon}, those common variances whose values are greater than $50\%$ have been bolded. Bold values indicate that a certain factor reconstructs most of the variance of a given attribute. In other words, a given attribute is similar to a certain factor. From the point of view of attribute clustering, it is only interesting if more than one attribute is similar to a factor. If there are two or more attributes, which are similar to a factor, it can be said that each of these attributes belongs to the same cluster.
In the table analyzed, it can be seen that attributes \textit{A2>3}, \textit{A3>1} and \textit{A3>2} are similar to the factor $F1$, and therefore belong to one cluster. It can also be seen that attributes \textit{A1>3}, \textit{A2>1} and \textit{A2>2} are similar to the factor $F2$.
So these three attributes also form one cluster. Since the other bold values of the common variance are single-attribute clusters, they are therefore not interesting from the point of view of attribute clustering.

\begin{table}[h!]
\centering
\caption{\textit{Simple Weather Forecast} dataset -- final factor loading matrix after rotation}\label{weatherFinFactRotated}
\fontsize{8.5}{12}\selectfont{
	\begin{tabular}{|c|c|c|c|c|c|c|} 	\hline
		&$F1$&$F2$&$F3$&$F4$&Common variances\\ \hline
		\textit{A1>1}&-0.063&0.340&0.776&0.060&72.5\%\\
		\textit{A1>2}&-0.069&0.515&-0.811&-0.194&96.5\%\\
		\textit{A1>3}&0.128&-0.826&-0.011&0.123&71.3\%\\
		\textit{A2>1}&-0.292&0.804&0.054&0.253&79.9\%\\
		\textit{A2>2}&-0.525&-0.692&-0.083&-0.091&77.0\%\\
		\textit{A2>3}&0.868&-0.046&0.037&-0.154&78.0\%\\
		\textit{A3>1}&-0.918&0.105&0.173&-0.067&88.8\%\\
		\textit{A3>2}&0.918&-0.105&-0.173&0.067&88.8\%\\
		\textit{A4}&-0.023&0.058&-0.026&0.966&93.8\%\\
		\textit{A5}&0.296&-0.034&-0.757&0.303&75.4\%
		\\ \hline
	\end{tabular}
}
\end{table}

\begin{table}[h!]
\centering
\caption{\textit{Simple Weather Forecast} dataset -- final matrix of common variances after rotation}\label{weatherFinCommonRotated}
\fontsize{8.5}{12}\selectfont{
	\begin{tabular}{|c|c|c|c|c|c|c|} 	\hline
		&$F1$&$F2$&$F3$&$F4$&Common variances\\ \hline
		\textit{A1>1}&0.40\%&11.57\%&\textbf{60.19\%}&0.36\%&72.5\%\\
		\textit{A1>2}&0.47\%&26.50\%&\textbf{65.80\%}&3.75\%&96.5\%\\
		\textit{A1>3}&1.63\%&\textbf{68.15\%}&0.01\%&1.50\%&71.3\%\\
		\textit{A2>1}&8.53\%&\textbf{64.64\%}&0.29\%&6.40\%&79.9\%\\
		\textit{A2>2}&27.60\%&47.91\%&0.69\%&0.82\%&77.0\%\\
		\textit{A2>3}&\textbf{75.26\%}&0.21\%&0.14\%&2.36\%&78.0\%\\
		\textit{A3>1}&\textbf{84.30\%}&1.11\%&2.98\%&0.44\%&88.8\%\\
		\textit{A3>2}&\textbf{84.30\%}&1.11\%&2.98\%&0.44\%&88.8\%\\
		\textit{A4}&0.05\%&0.34\%&0.07\%&\textbf{93.31\%}&93.8\%\\
		\textit{A5}&8.78\%&0.12\%&\textbf{57.33\%}&9.17\%&75.4\%
		\\ \hline
	\end{tabular}
}
\end{table}

\subsubsection{Rotation of factors}
Now the question arises whether Varimax rotation could change anything in the results of attribute clustering. To answer this question, a rotation of the factors in Table \ref{weatherFinFact} was performed.
After rotation, a new factor loadings matrix was obtained (Table \ref{weatherFinFactRotated}), as well as a new common variance matrix (Table \ref{weatherFinCommonRotated}).

An analysis of the bolded values in the common variances in Table \ref{weatherFinCommonRotated} shows that there is another cluster that contains three attributes similar to the $F3$ factor.
On the other hand, in the cluster of attributes similar to the factor $F2$, the attribute \textit{A2>2} is no longer absolutely (in the sense of the absolute majority variance rule) similar to this factor.
Despite this, the similarity of $47.91\%$ indicates a significant influence of factor two on attribute \textit{A2>2}. 
This influence accounts for more than half of the total variance ($77.0\%$) explained by the four factors. 
This means that, using the weaker relative majority of variance rule, the attribute \textit{A2>2} can also be considered as belonging to the cluster of attributes similar to the factor $F2$.

\begin{table}[h!]
\centering
\caption{\textit{Simple Weather Forecast} dataset -- minimized common variance matrix for the rule of absolute majority of variance}\label{weatherFinCommonRotatedAbs}
\fontsize{8.5}{12}\selectfont{
	\begin{tabular}{|c|c|c|c|c|c|} 	\hline
		&$F1$&$F2$&$F3$&Common variances\\ \hline
		\textit{A1>1}&0.4\%&11.6\%&\textbf{60.2\%}&72.5\%\\
		\textit{A1>2}&0.5\%&26.5\%&\textbf{65.8\%}&96.5\%\\
		\textit{A1>3}&1.6\%&\textbf{68.1\%}&0.0\%&71.3\%\\
		\textit{A2>1}&8.5\%&\textbf{64.6\%}&0.3\%&79.9\%\\
		\textit{A2>3}&\textbf{75.3\%}&0.2\%&0.1\%&78.0\%\\
		\textit{A3>1}&\textbf{84.3\%}&1.1\%&3.0\%&88.8\%\\
		\textit{A3>2}&\textbf{84.3\%}&1.1\%&3.0\%&88.8\%\\
		\textit{A5}&8.8\%&0.1\%&\textbf{57.3\%}&75.4\%
		\\ \hline
	\end{tabular}
}
\end{table}

\begin{figure}[h!]
\centering
\includegraphics[width=0.60\textwidth]{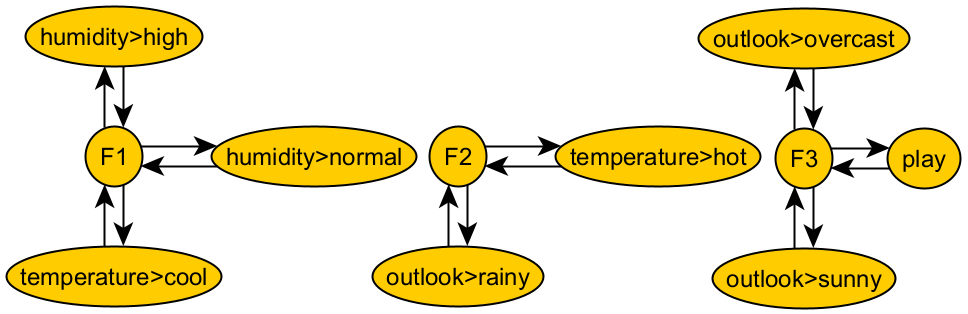}
\caption{Absolute similarity of attributes to factors for encoded \textit{Simple Weather Forecast} dataset}\label{weatherAbs}
\end{figure}

\begin{table}[h!]
\centering
\caption{\textit{Simple Weather Forecast} dataset -- minimized common variance matrix for the rule of relative majority of variance}\label{weatherFinCommonRotatedRelat}
\fontsize{8.5}{12}\selectfont{
	\begin{tabular}{|c|c|c|c|c|c|} 	\hline
		&$F1$&$F2$&$F3$&Common variances\\ \hline
		\textit{A1>1}&0.4\%&11.6\%&\textbf{60.2\%}&72.5\%\\
		\textit{A1>2}&0.5\%&26.5\%&\textbf{65.8\%}&96.5\%\\
		\textit{A1>3}&1.6\%&\textbf{68.1\%}&0.0\%&71.3\%\\
		\textit{A2>1}&8.5\%&\textbf{64.6\%}&0.3\%&79.9\%\\
		\textit{A2>2}&27.6\%&\textbf{47.9\%}&0.7\%&77.0\%\\
		\textit{A2>3}&\textbf{75.3\%}&0.2\%&0.1\%&78.0\%\\
		\textit{A3>1}&\textbf{84.3\%}&1.1\%&3.0\%&88.8\%\\
		\textit{A3>2}&\textbf{84.3\%}&1.1\%&3.0\%&88.8\%\\
		\textit{A5}&8.8\%&0.1\%&\textbf{57.3\%}&75.4\%
		\\ \hline
	\end{tabular}
}
\end{table}

\begin{figure}[h!]
\centering
\includegraphics[width=0.60\textwidth]{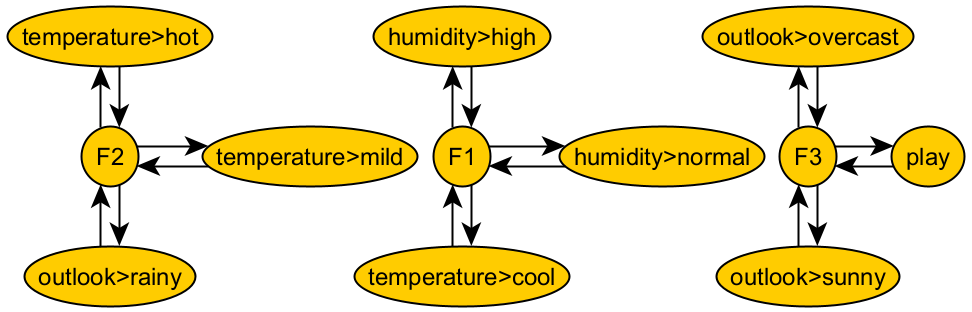}
\caption{Relative similarity of attributes to factors for encoded dataset \textit{Simple Weather Forecast}}\label{weatherRelat}
\end{figure}

\subsubsection{Final clustering of random attributes}
The matrix of common variances obtained after rotation (Table \ref{weatherFinCommonRotated}) provides information on the strength of the correlation relationship between attributes and factors obtained for a specific factor model. 
On the basis of this matrix, it is possible to proceed to clustering attributes that are similar to individual factors. From the point of view of clustering attributes, it is possible to extract both relevant and redundant information:
\begin{itemize}[nosep]
\item Relevant information is that which proves that more than one attribute is similar to a factor.
\item Redundant information is that which proves that at most one attribute is similar to a factor.
\end{itemize}
Thus, in Table \ref{weatherFinCommonRotated}, it can be seen that only attribute \textit{A4} is similar to factor $F4$. 
A cluster containing attributes similar to the $F4$ factor contains only one attribute in it. Therefore, it is a trivial cluster, and therefore this cluster is not interesting from the point of view of attribute clustering. This means that from the point of view of attribute clustering, both the \textit{A4} attribute and the $F4$ factor are redundant. Therefore, attribute \textit{A4} and factor $F4$ can be omitted from the clustering procedure.

On the other hand, if no factor reproduces at least a relative majority of the variance of any attribute, that factor should also be considered redundant from the point of view of attribute clustering. In view of this, it can also be omitted from the attribute clustering procedure.
After eliminating trivial clusters and factors that do not reproduce at least a relative majority of the variance of any attribute, clustering can be implemented using one of two rules:
\begin{enumerate}[nosep]
\item The absolute majority of variance rule should be applied.
\item The rule of relative majority of variance can be applied.
\end{enumerate}
From the point of view of clustering according to the absolute majority of variance, it can be seen that in the matrix shown in Table \ref{weatherFinCommonRotated}, the attribute \textit{A2>2} can also be reduced, since no factor represents the majority (more than half) of its variance.
After reducing the factor $F4$, as well as the attributes \textit{A2>2} and \textit{A4}, what remains is a matrix of common variances (Table \ref{weatherFinCommonRotatedAbs}), from which the clusters obtained according to the absolute majority of variance criterion can be extracted.
Thus, it can be noted that:
\begin{itemize}[nosep]
\item The first cluster is formed by attributes similar to the $F1$ factor. There are attributes \textit{A2>3}, \textit{A3>1} and attribute \textit{A3>2}.
\item The second cluster is formed by attributes similar to the $F2$ factor. This cluster includes attributes \textit{A1>3} and \textit{A2>1}.
\item The third cluster is formed by attributes similar to the $F3$ factor. Here are attributes \textit{A1>1}, \textit{A1>2} and \textit{A5}.
\end{itemize}
All the above clusters, obtained according to the total majority of variance rule, are shown in Figure \ref{weatherAbs} as connected components of the graph.
On the other hand, it can be seen that according to the relative majority of variance rule, the cluster of attributes similar to the factor $F2$ also includes the attribute \textit{A2>2}.
Despite the fact that the common variance of the \textit{A2>2} attribute and the $F2$ factor is $47.9\%$ (that is, it does not exceed the $50\%$ level), this variance accounts for more than half ($77\%$) of the total variance of the \textit{A2>2} attribute represented by the selected four factors. Clusters obtained according to the rule of relative majority of variance can be extracted from Table \ref{weatherFinCommonRotatedRelat}.
For this rule, the cluster of attributes similar to the $F2$ factor contains three attributes: \textit{A1>3}, \textit{A2>1} and \textit{A2>2}. 
All clusters obtained using the relative majority variance rule are shown as a graph in Figure \ref{weatherRelat}. 

Comparing the attribute clustering results obtained according to the absolute majority variance rule with the clustering results obtained according to the relative majority variance rule, it can be seen that the weakening of the classification rule resulted in the fact that one additional attribute arrived in the cluster of attributes similar to the \textit{F2} factor. This attribute represents the \textit{temperature>mild} class.

\begin{table}[h!]
\centering
\caption{Naming schemes applied to attributes in Mashroom dataset}\label{mashCodComp}
\fontsize{8.5}{12}\selectfont{
	\begin{tabular}{|c|c|c|} 	\hline
		No.&Full name of the encoded attribute&Short name of the encoded attribute \\ \hline
		1.&class&A1 \\
		2.&cap-shape&A2 \\
		3.&cap-surface&A3 \\
		4.&cap-color>n&A4>1 \\
		5.&cap-color>b&A4>2 \\
		6.&cap-color>c&A4>3 \\
		7.&cap-color>g&A4>4 \\
		8.&cap-color>r&A4>5 \\
		9.&cap-color>p&A4>6 \\
		10.&cap-color>u&A4>7 \\
		11.&cap-color>e&A4>8 \\
		12.&cap-color>w&A4>9 \\
		13.&cap-color>y&A4>10 \\
		14.&bruises?&A5 \\
		15.&odor&A6 \\
		16.&gill-attachment&A7 \\
		17.&gill-spacing&A8 \\
		18.&gill-size&A9 \\
		19.&gill-color&A10 \\
		20.&stalk-shape&A11 \\
		21.&stalk-surface-above-ring&A12 \\
		22.&stalk-surface-below-ring&A13 \\
		23.&stalk-color-above-ring&A14 \\
		24.&stalk-color-below-ring&A15 \\
		25.&veil-color&A16 \\
		26.&ring-number&A17 \\
		27.&ring-type&A18 \\
		28.&spore-print-color&A19 \\
		29.&population&A20 \\
		30.&habitat&A21 
		\\
		\hline
	\end{tabular}
}
\end{table}

\begin{table}[h!]
\centering
\caption{Percentage of variance explained by successive factors for \textit{Mushroom} dataset}\label{mushAverCriterium}
\fontsize{8.5}{12}\selectfont{
	\begin{tabular}{c|c|c|c|c} \hline 
		Factor & \multirow{2}{*}{Eigenvalue} & Cumulative & Percentage of the variance & Cumulative percentage  \\ 
		no. & & eigenvalues & explained by each factor &  of explained variance \\ \hline 
		$\vdots$&$\vdots$&$\vdots$&$\vdots$&$\vdots$\\
		8&1.19&18.47&3.98\%&61.57\%\\
		9&1.15&19.62&3.83\%&65.40\%\\
		10&1.10&20.72&3.68\%&69.08\%\\
		11&1.00&21.73&3.34\%&72.42\%\\
		12&0.99&22.71&3.29\%&75.72\%\\
		13&0.98&23.69&3.26\%&78.98\%\\
		14&0.81&24.50&2.70\%&81.68\%\\
		15&0.77&25.28&2.58\%&84.26\%\\
		$\vdots$&$\vdots$&$\vdots$&$\vdots$&$\vdots$
		\\ \hline
	\end{tabular}
}
\end{table}

\begin{table}[h!]
\centering
\caption{\textit{Mushroom} dataset -- variances represented by several successive factors: ScreePlt, MinVar and AverVar}\label{mushMinVar}
\fontsize{8.5}{12}\selectfont{
	\begin{tabular}{|c|c|c|c|c|c|c|c|c|c|c|} 	\hline
		Factor no.&$\cdots$&8&9&10&11&12&13&14&15&$\cdots$\\ \hline
		ScreePlt&$\cdots$&4.0\%&3.8\%&3.7\%&3.3\%&3.3\%&3.3\%&2.7\%&2.6\%&$\cdots$\\
		MinVar&$\cdots$&9.5\%&16.6\%&19.8\%&19.8\%&46.8\%&53.8\%&58.6\%&60.5\%&$\cdots$\\
		AverVar&$\cdots$&61.6\%&65.4\%&69.1\%&72.4\%&75.7\%&79.0\%&81.7\%&84.3\%&$\cdots$\\
		MinVarID&$\cdots$&\textit{A4>7}&\textit{A4>7}&\textit{A4>3}&\textit{A4>3}&\textit{A2}&\textit{A3}&\textit{A21}&\textit{A10}&$\cdots$
		\\
		\hline
	\end{tabular}
}
\end{table}

\begin{figure}[!t]
\centering
\includegraphics[width=0.75\textwidth]{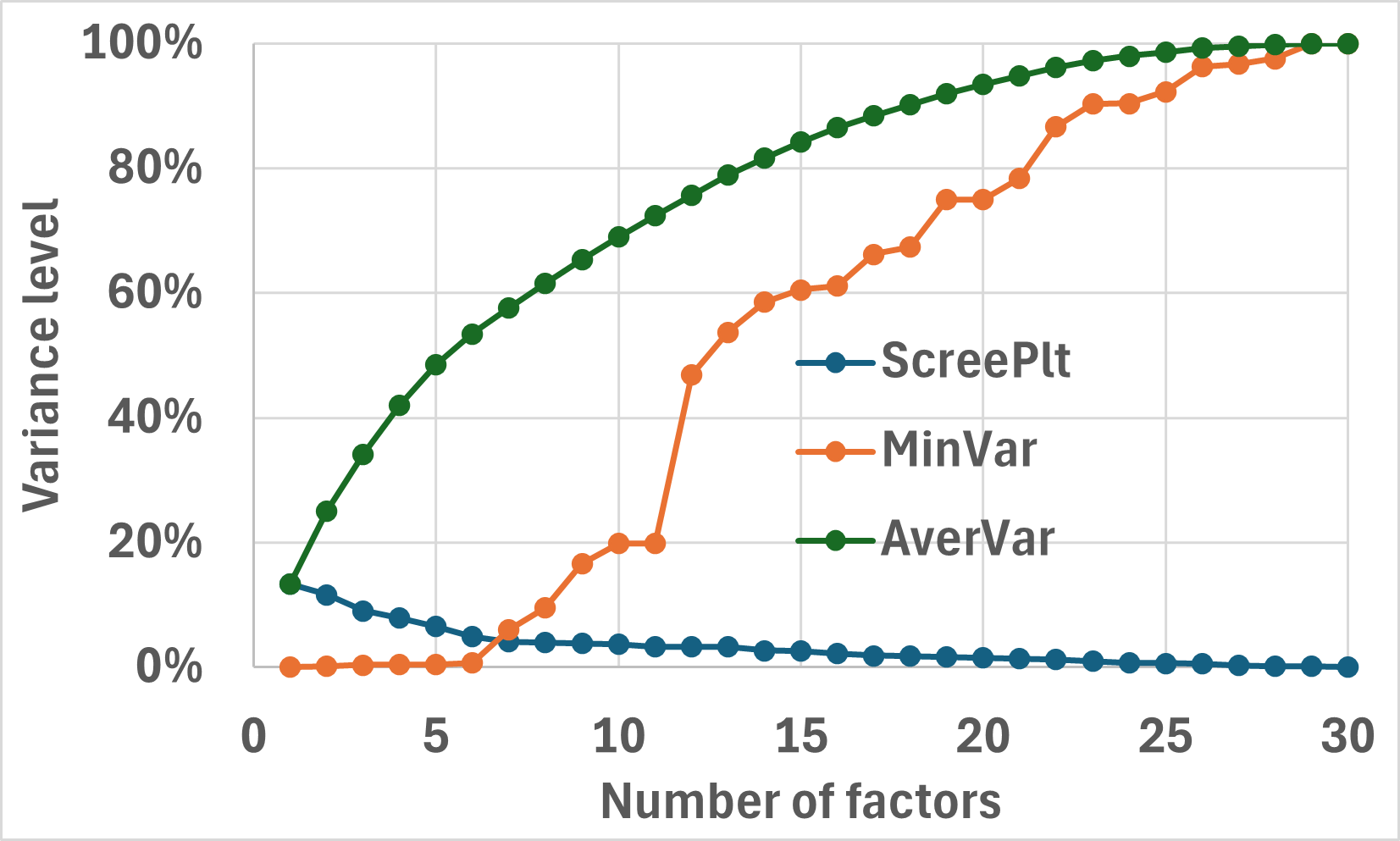}
\caption{\textit{Mushroom} dataset -- variances represented by several successive factors: ScreePlt, MinVar and AverVar}\label{mushMinVarPlot}
\end{figure}

\begin{figure}[h!]
\centering
\includegraphics[width=0.50\textwidth]{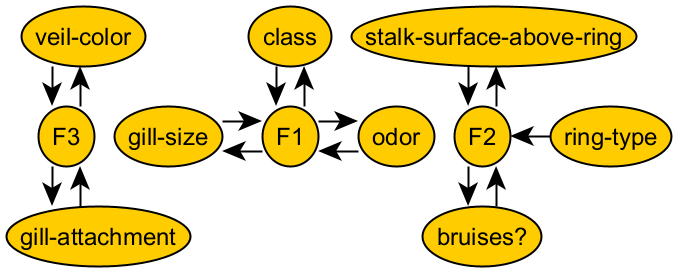}
\caption{Absolute similarity of attributes to factors for the encoded \textit{Mushroom} dataset}\label{mushAbs}
\end{figure}

\begin{figure}[h!]
\centering
\includegraphics[width=0.75\textwidth]{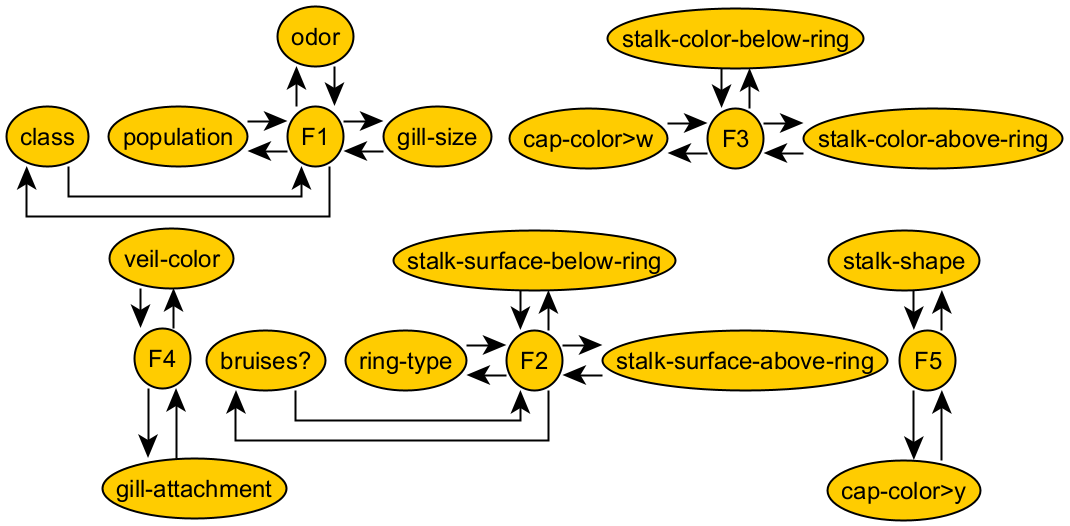}
\caption{Relative similarity of attributes to factors  for the encoded \textit{Mushroom} dataset}\label{mushPlur}
\end{figure}

\begin{figure}[h!]
\centering
\includegraphics[width=0.65\textwidth]{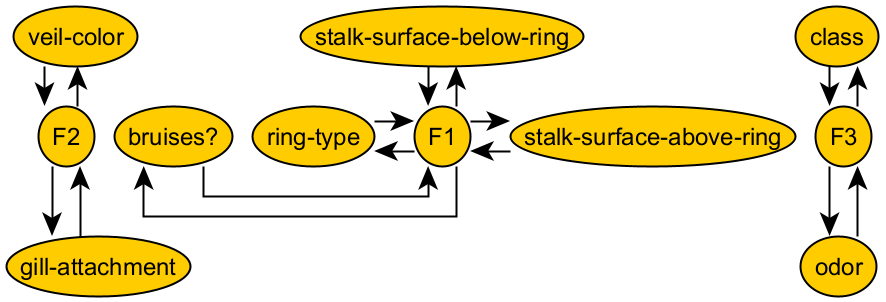}
\caption{Absolute similarity of attributes to factors for the ranks of the numerically encoded \textit{Mushroom} dataset}\label{mush2Major}
\end{figure}

\begin{figure}[h!]
\centering
\includegraphics[width=0.50\textwidth]{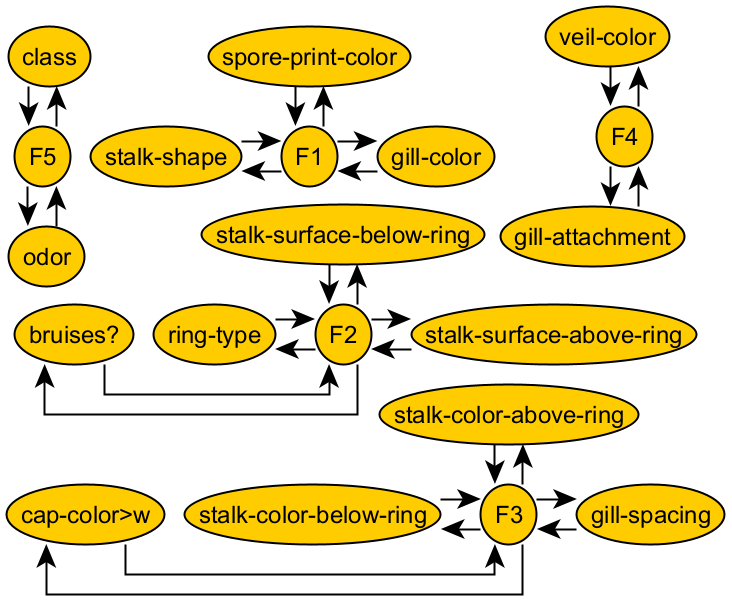}
\caption{Relative similarity of attributes to factors for the ranks of the numerically encoded \textit{Mushroom} dataset}\label{mush2Plur}
\end{figure}

\subsection{Dataset No. 2 -- \textit{Mushroom}}
Another study used a dataset called \textit{Mushroom} \cite{mushroom1981}, which contains descriptions of 23 species of mushrooms from the families \textit{Agaricus} and \textit{Lepiota}.  Each species was classified as edible, poisonous, or of unknown edibility.  In the dataset used here, the last class was classified as poisonous.

The \textit{Mushroom} dataset contains nominal data describing $8124$ points in the space of 23 attributes. Before attribute clustering, $2$ columns containing veil-type and stalk-root attribute information were arbitrarily removed from the original dataset:
\begin{itemize}[nosep]
\item The veil-type attribute took only one value equal to ''p''.
\item The stalk-root attribute was missing $2480$ values.
\end{itemize}
The decision on the first attribute above was obvious. A fixed value for an attribute does not carry any interesting information. In the case of the second attribute, there was no rationale for filling in the missing values in any way. Therefore, two options were considered: delete $2480$ rows with missing values or delete one column. The decision was made to remove the column. As a result, a dataset containing $21$ attributes was analyzed.

In the case of the cap-color attribute, the attribute was found to contain 10 different value classes. The classes containing the values “r” and “u” are equicardinal classes containing 16 elements each. Therefore, this attribute was encoded using one-hot encoding methods. Consequently, the single column corresponding to the cap-color attribute was replaced by $10$ binary columns. After encoding, the analyzed dataset contained $30$ columns. Table \ref{mashCodComp} shows the attribute naming system used in the analysis.
\subsubsection{Clustering of encoded attributes}
After the attributes were encoded, the standard attribute clustering procedure was implemented, just as described in subsection \ref{Dat1}. 
First, a factor analysis was performed. In it, an appropriate number of factors were selected and Varimax rotation was performed. The attributes were then clustered, both according to the absolute majority variance rule and according to the relative majority variance rule.

Table \ref{mushAverCriterium} shows part of the table with the values of (average) variance that are explained by successive factors. 
Assuming that a factor model should explain at least $70\%$ of the average variance of all modeled attributes, it can be read in Table \ref{mushAverCriterium} that a model with $11$ factors is sufficient for this purpose.  
On the other hand, it can be read in Table $23$ that although the average variance is reproduced satisfactorily by $11$ factors, these $11$ factors are enough to reproduce only $19.8\%$ of the variance of the \textit{A4>3} attribute.
Analyzing Table \ref{mushMinVar} and Figure \ref{mushMinVarPlot}, it can be seen that $14$ factors are needed to reproduce most of the variance of each individual attribute.

Varimax rotation was performed for the model consisting of $14$ factors. Then, using Algorithm \ref{algor2}, attribute clustering was performed based on the common variance matrix. The clustering results obtained using the absolute majority variance rule are shown in Figure \ref{mushAbs}, and the result of attribute clustering using the relative majority variance rule is shown in Figure \ref{mushPlur}.

\subsubsection{Effect of ranking of encoded attributes on clustering results}
The analyzed dataset consists only of nominal attributes. Due to the existing equicardinal value classes, one attribute was encoded using the one-hot method. The remaining attributes were encoded with class cardinalities. Data encoded with class cardinalities can be ranked. After they were ranked, the attributes were clustered again. Figure \ref{mush2Major} shows the results of attribute clustering obtained according to the absolute majority of variance rule. Figure \ref{mush2Plur} shows the clustering results obtained according to the relative majority of variance rule. On the one hand, attribute ranking did not change the number of factors needed to represent the majority of the variance of each attribute. On the other hand, comparing the results of attribute clustering before ranking (Figure \ref{mushAbs} and Figure \ref{mushPlur}) and after ranking (Figure \ref{mush2Major} and Figure \ref{mush2Plur}), it was found that some changes in the results of attribute clustering appeared with the ranking of encoded attributes.

\begin{table}[h!]
\centering
\caption{Naming schemes applied to attributes in \textit{Automobile} dataset}\label{autoCodComp}
\fontsize{8.5}{12}\selectfont{
	\begin{tabular}{c|c|c||c|c|c} \hline 
		\multirow{2}{*}{No.} &Full name of the &Short name of the&\multirow{2}{*}{No.} &Full name of the &Short name of the\\
		&coded attribute&coded attribute&&coded attribute&coded attribute\\ \hline
		1.&symboling&A1&27.&num-of-doors&A6 \\
		2.&normalized-losses&A2&28.&body-style&A7 \\
		3.&make>alfa-romero&A3>1&29.&drive-wheels&A8 \\
		4.&make>audi&A3>2&30.&wheel-base&A9 \\
		5.&make>bmw&A3>3&31.&length&A10 \\
		6.&make>chevrolet&A3>4&32.&width&A11 \\
		7.&make>dodge&A3>5&33.&height&A12 \\
		8.&make>honda&A3>6&34.&curb-weight&A13 \\
		9.&make>isuzu&A3>7&35.&engine-type>dohc&A14>1 \\
		10.&make>jaguar&A3>8&36.&engine-type>dohcv&A14>2 \\
		11.&make>mazda&A3>9&37.&engine-type>l&A14>3 \\
		12.&make>mercedes-benz&A3>10&38.&engine-type>ohc&A14>4 \\
		13.&make>mercury&A3>11&39.&engine-type>ohcf&A14>5 \\
		14.&make>mitsubishi&A3>12&40.&engine-type>ohcv&A14>6 \\
		15.&make>nissan&A3>13&41.&engine-type>rotor&A14>7 \\
		16.&make>peugot&A3>14&42.&num-of-cylinders&A15 \\
		17.&make>plymouth&A3>15&43.&engine-size&A16 \\
		18.&make>porsche&A3>16&44.&fuel-system&A17 \\
		19.&make>renault&A3>17&45.&bore&A18 \\
		20.&make>saab&A3>18&46.&stroke&A19 \\
		21.&make>subaru&A3>19&47.&compression-ratio&A20 \\
		22.&make>toyota&A3>20&48.&horsepower&A21 \\
		23.&make>volkswagen&A3>21&49.&peak-rpm&A22 \\
		24.&make>volvo&A3>22&50.&city-mpg&A23 \\
		25.&fuel-type&A4&51.&highway-mpg&A24 \\
		26.&aspiration&A5&52.&price&A25 \\ \hline
	\end{tabular}
}
\end{table}

\begin{table}[h!]
\centering
\caption{Attribute variances from the \textit{Automobile} dataset, represented by successive factors : ScreePlt, MinVar and AverVar}\label{autoMinVar}
\fontsize{8.5}{12}\selectfont{
	\begin{tabular}{|c|c|c|c|c|c|c|c|c|c|} 	\hline
		Factor no.&$\ldots$&14&15&16&$\ldots$&22&23&24&$\ldots$\\ \hline
		ScreePlt&$\ldots$&2.18\%&2.15\%&2.04\%&$\ldots$&1.92\%&1.92\%&1.76\%&$\ldots$ \\
		MinVar&$\ldots$&0\%&0\%&0\%&$\ldots$&0\%&64.31\%&67.91\%&$\ldots$ \\
		AverVar&$\ldots$&69.00\%&71.15\%&73.19\%&$\ldots$&84.82\%&86.74\%&88.51\%&$\ldots$ \\
		MinVarID&$\ldots$&A3>1&A3>1&A3>1&$\ldots$&A14>7&A3>16&A5&$\ldots$			
		\\
		\hline
	\end{tabular}
}
\end{table}

\begin{figure}[h!]
\centering
\includegraphics[width=0.75\textwidth]{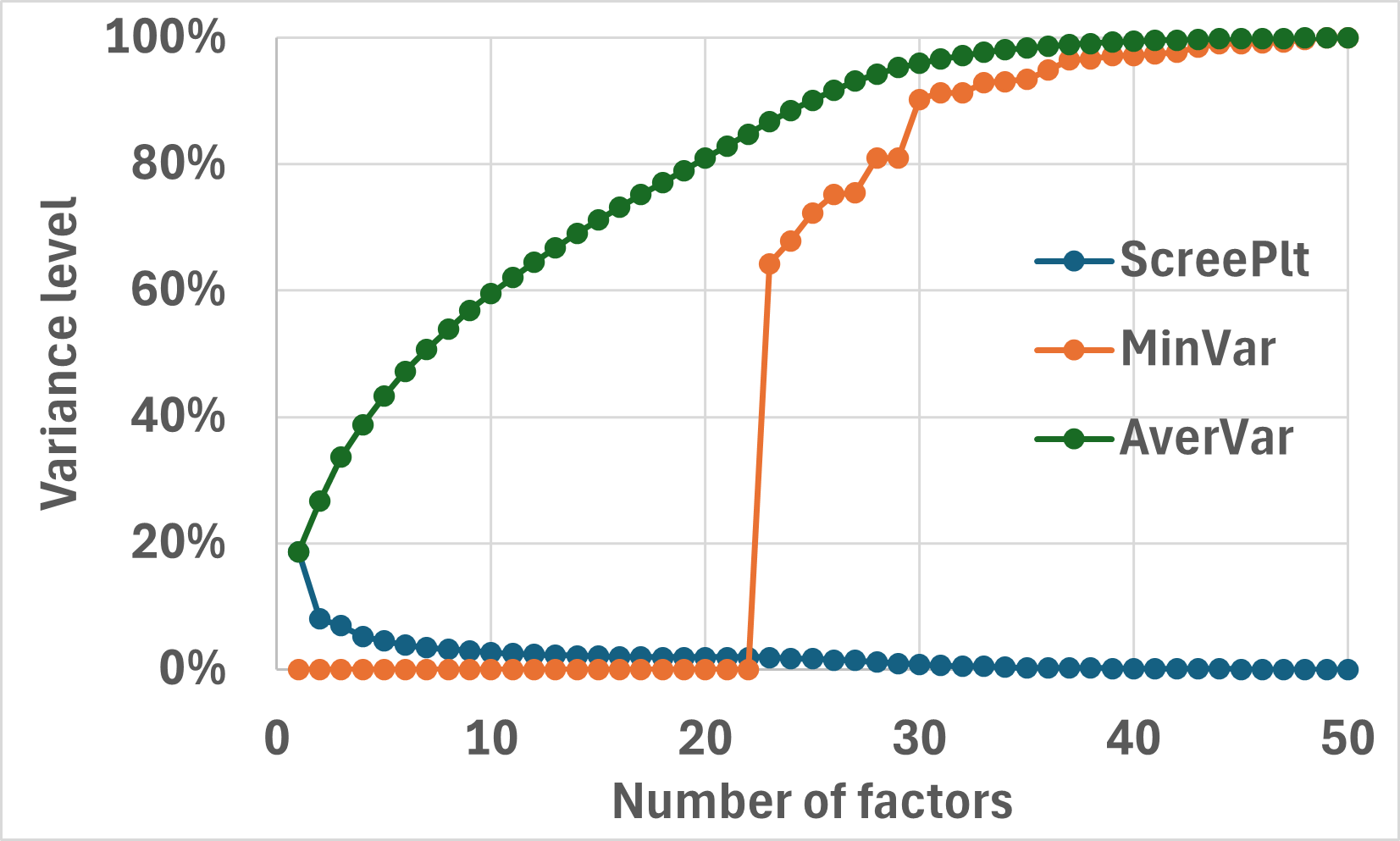}
\caption{\textit{Automobile} dataset -- attribute variances: minimum variance (MinVar) and average variance (AverVar), reconstructed by successive factors, shown against a normalized scree plot (ScreePlt).}\label{autoMinVarPlot}
\end{figure}

\begin{figure}[h!]
\centering
\includegraphics[width=0.90\textwidth]{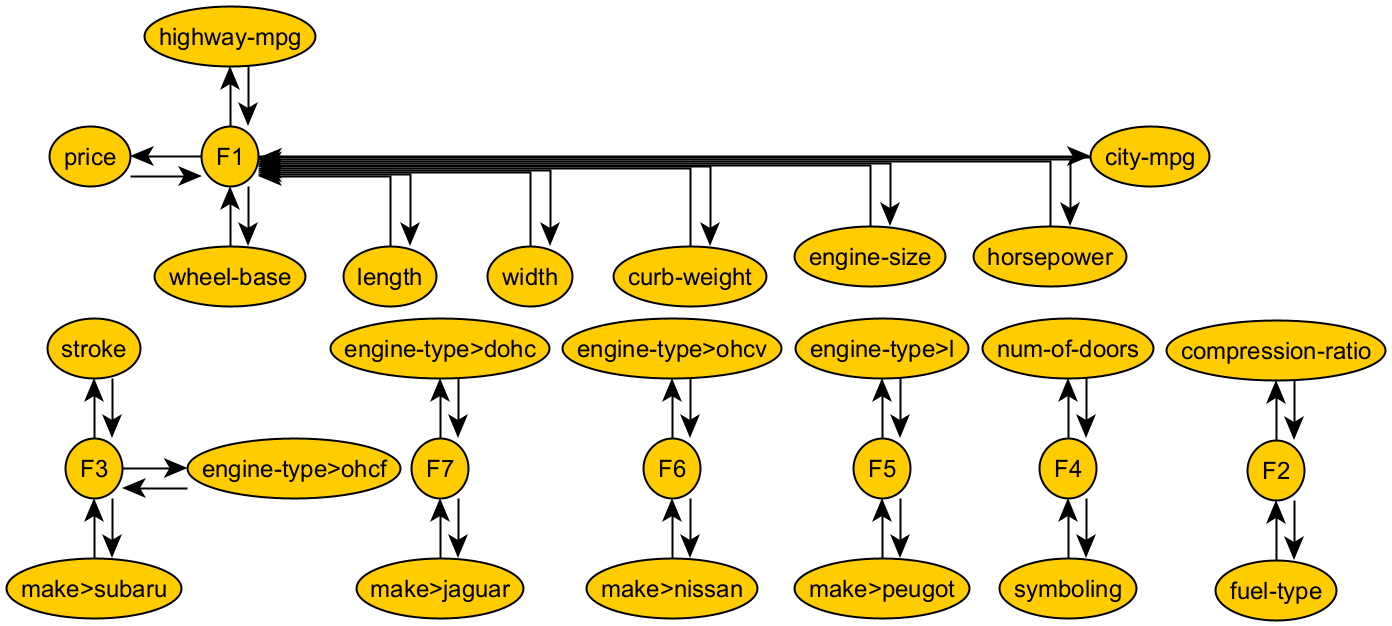}
\caption{Absolute similarity of attributes to factors for the encoded \textit{Automobile} dataset}\label{auto0Abs}
\end{figure}

\begin{figure}[h!]
\centering
\includegraphics[width=0.95\textwidth]{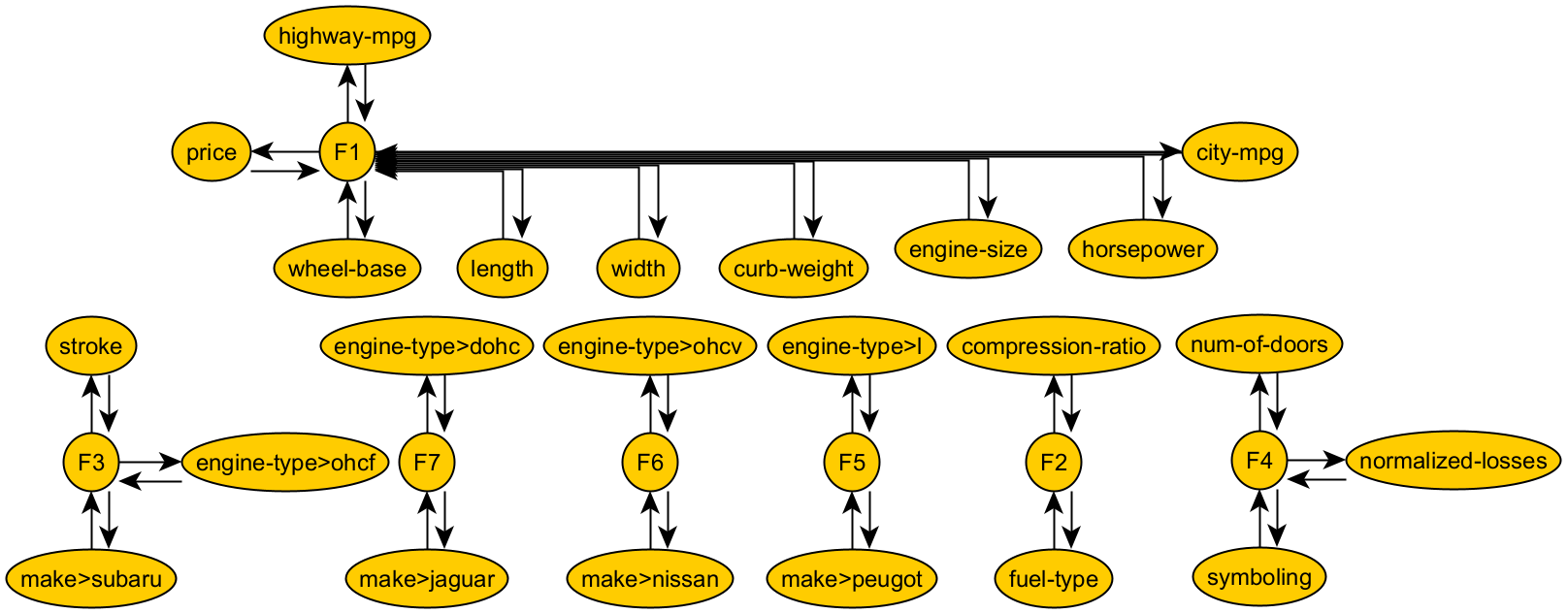}
\caption{Relative similarity of attributes to factors  for the encoded \textit{Automobile} dataset}\label{auto0Plur}
\end{figure}

\begin{figure}[h!]
\centering
\includegraphics[width=0.90\textwidth]{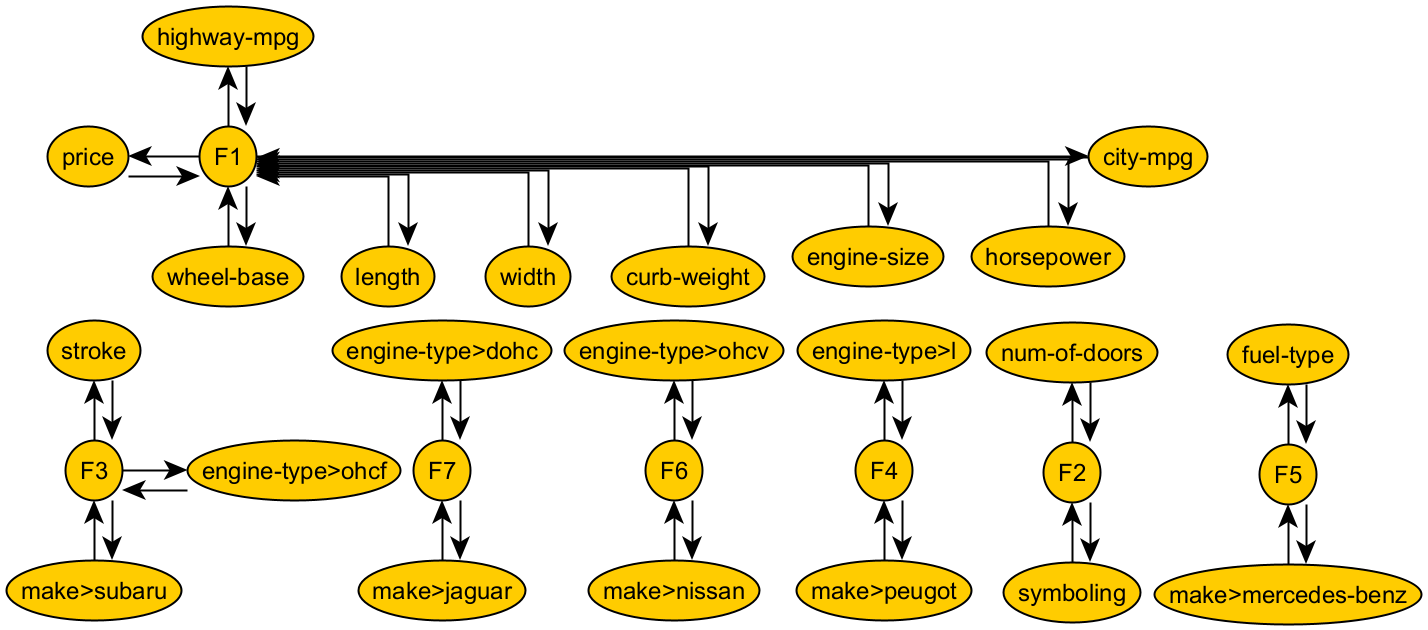}
\caption{Absolute similarity of attributes to factors for the ranks of  numerically encoded \textit{Automobile} dataset}\label{auto2Major}
\end{figure}

\begin{figure}[h!]
\centering
\includegraphics[width=0.90\textwidth]{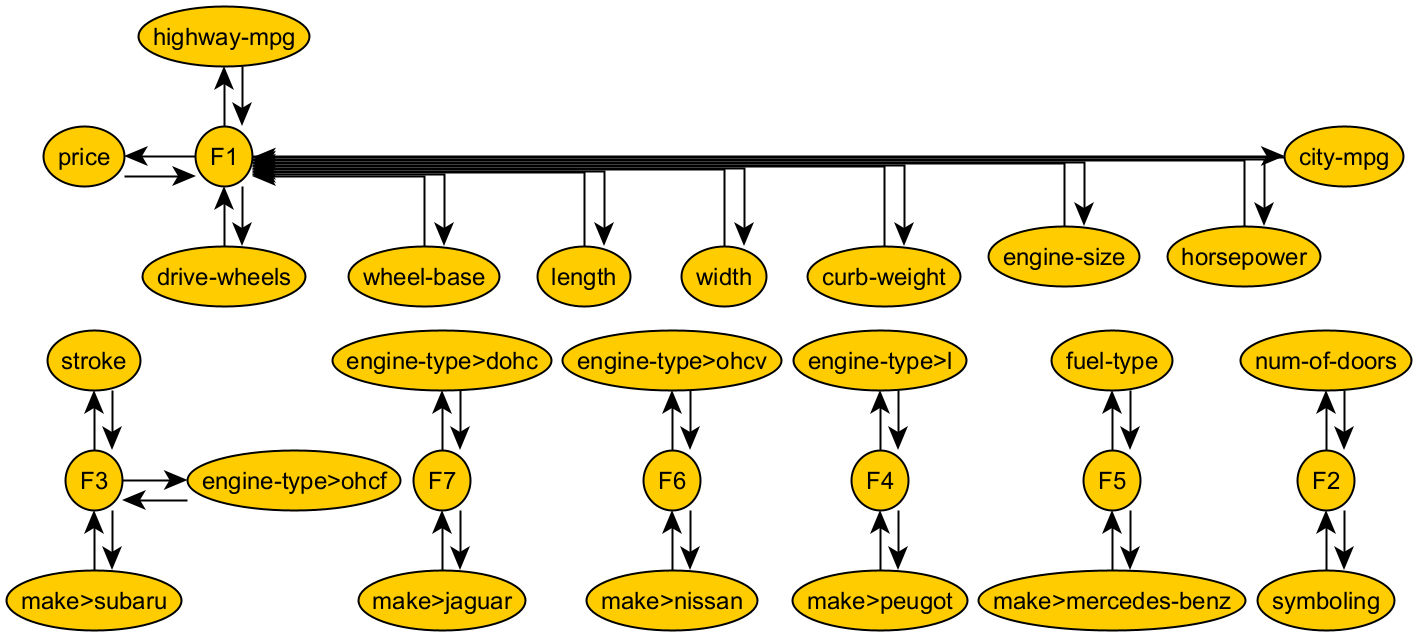}
\caption{Relative similarity of attributes to factors for the ranks of  numerically encoded \textit{Automobile} dataset}\label{auto2Plur}
\end{figure}

\subsection{Dataset No. 3 -- \textit{Automobile}}
The following test of the attribute clustering algorithm uses a dataset named \textit{Automobile} \cite{automobile1985}. 
The \textit{Automobile} dataset is contained in a table with $26$ columns and $205$ rows. Each column represents one attribute. Each attribute is measured on different scales.

The original dataset was assumed to describe $15$ continuous attributes, $10$ nominal attributes and $1$ numerical attribute (''\textit{symboling}'').  Two attributes that are essentially numerical attributes  (\textit{number-of-doors=\{two,four\}}, \textit{number-of-cylinders=\{three,four,five,six,eight\}}) were classified as nominal attributes. Before starting for analysis, their qualification was changed, assuming that they are numeric attributes: \textit{num-of-doors=\{2,4\}}, \textit{num-of-cylinders=\{3,4,5,6,8\}}.
The dataset also contains missing values, which are marked as ''\textit{?}''. Attribute number $2$ contains $41$ missing values, attributes numbered $6$, $22$ and $23$ each contain $2$ missing values, and attributes numbered $19$, $20$ and $26$ each contain $4$ missing values. Before clustering, it was decided to reject all rows with missing attribute values. This left $159$ rows from the original data table for further analysis.
After reducing the rows containing the missing values, it was found that the \textit{engine-location} attribute takes only a single value (\textit{front}). This single value does not contain any interesting information. Therefore, it was decided to ignore this attribute in further analysis.
After all the changes made, $25$ attributes remain to be analyzed. Now the data table contains $25$ columns and $159$ rows.

Assuming nomenclature as in subsection \ref{numVSnom}, it was assumed that the data contains $18$ numerical attributes and $7$ nominal attributes.
Among these $7$ nominal attributes, $2$ attributes (\textit{make} and \textit{engine-type}) contain equicardinal classes. Thus, $5$ nominal attributes were encoded with class cardinalities, while $2$ attributes containing equicardinal classes were encoded with the one-hot method. Finally, after encoding, the data contains $52$ columns and $159$ rows. Table \ref{autoCodComp} shows the names of the encoded attributes.

Table \ref{autoMinVar} shows the partial distribution of the variance of attributes reproduced by the factors for the \textit{Automobile} dataset. From this table it can be read that a model with $15$ factors is sufficient to reproduce $70\%$ of the average variance of all attributes by a factor model. However, with $15$ factors, the variances of some attributes are not sufficiently represented (e.g. \textit{A3>1}). On the other hand, it can be read from Table \ref{autoMinVar} and Figure \ref{autoMinVarPlot} that $23$ factors are needed to reproduce most of the variance of each attribute.

After selecting the right number of factors and rotating them, attribute clustering was performed. Figure \ref{auto0Abs} shows the result of clustering according to the absolute majority variance rule, while Figure \ref{auto0Plur} shows the results of clustering according to the relative majority variance rule.

The data analyzed contained two types of attributes. On the one hand, there were nominal attributes, which were either one-hot encoded or were encoded with cardinalities, On the other hand, there were numeric attributes in the dataset. Both cardinality-coded attributes and numeric attributes were ranked. After ranking, the attributes were clustered. Figure \ref{auto2Major} shows the result of clustering according to the absolute majority variance rule, while Figure \ref{auto2Plur} shows the results of clustering according to the relative majority variance rule.

\begin{figure}[h!]
\centering
\includegraphics[width=0.30\textwidth]{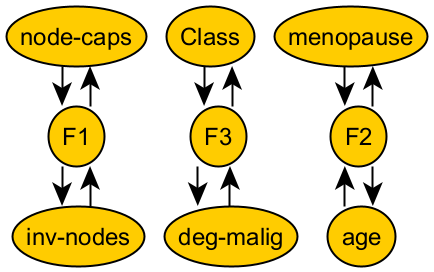}
\caption{Similarity between attributes and factors (absolute and relative) for encoded \textit{Breast Cancer} dataset}\label{breastAbs}
\end{figure}

\subsection{Dataset No. 4 -- \textit{Breast Cancer}}
Another dataset used to test the attribute clustering algorithm is the \textit{Breast Cancer} dataset \cite{breastCancer1988}. In the original dataset\footnote{This breast cancer domain was obtained from the University Medical Centre,    Institute of Oncology, Ljubljana, Yugoslavia.  Thanks go to M. Zwitter and     M. Soklic for providing the data.}, the $10$ attributes observed $286$ times are presented as an array of size $286\times 10$. For two attributes, there are missing values marked as ''$?$''. There are $8$ missing values for the Node caps attribute, and one missing value for the Breast quadrant attribute. Before clustering, all rows with missing attribute values were discarded. Thus, $277$ rows from the initial data table were used for further analysis.
Among all $10$ attributes, $4$ attributes are ordinal:
\begin{itemize}[nosep]
\item The age of the patients was assigned to one of the following $5$-year ranges: $age=\{10-19, 20-29, 30-39, 40-49, 50-59, 60-69, 70-79, 80-89, 90-99\}$.
\item Tumor size was assigned to $11$ value ranges: $tumor-size=\{0-4, 5-9, 10-14, 15-19, 20-24, 25-29, 30-34, 35-39, 40-44, 45-49, 50-54, 55-59\}$.
\item The number of nodes with cancer was divided into $13$ ranges: $inv-nodes=\{0-2, 3-5, 6-8, 9-11, 12-14, 15-17, 18-20, 21-23, 24-26, 27-29, 30-32, 33-35, 36-39\}$.
\item The degree of cancer malignancy was graded on a $3$-point scale: $deg-malig=\{1, 2, 3\}$.
\end{itemize}
Due to the possibility of ordering the above $4$ attributes, the first $3$ attributes were encoded with successive integers. Since there is a relation $10-19<20-29<30-39<40-49<50-59<60-69<70-79<80-89<90-99$ in the set of values of the age attribute, therefore the \textit{age} attribute was encoded with a set of successive integers: $age=\{1, 2, 2, 4, 5, 6, 6, 7, 8, 9\}$. An analogous method was used for the \textit{tumor-size} attribute ($tumor-size=\{1, 2, 3, 4, 5, 6, 7, 8, 9, 10, 11, 12\}$), and for the \textit{inv-nodes} attribute ($inv-nodes=\{1, 2, 3, 4, 5, 6, 7, 8, 9, 10, 11, 12, 13\}$). The \textit{deg-malig} attribute was left unchanged ($deg-malig=\{1, 2, 3\}$). This encoding did not change the order among the values of the above attributes, and allowed them to be treated as numeric attributes.
On the other hand, the values of the remaining $6$ attributes (\textit{class}, \textit{menopause}, \textit{node-caps}, \textit{breast}, \textit{breast-quad}, \textit{irradiate}) could not be naturally ordered. Therefore, the values of these attributes were encoded with cardinalities of classes of identical values.

The solution to the problem of encoding attribute values presented above made it possible to treat them as numeric attributes, with all the consequences of this, including the possibility of ranking.
For the encoded data, first without ranking them, a clustering procedure was implemented, as in the previous three subsections. It was found that as many as $7$ factors were needed to represent most of the variance in each attribute (no less than $55\%$). Based on the factor model with $7$ factors, clusters were identified. 
For both the absolute majority variance rule and the relative majority variance rule, identical clusters were obtained.
Figure \ref{breastAbs} shows the clustering results.
Next, the clustering procedure was implemented for ranked data. In this case, it was found that 6 factors were needed to explain most of the variance in each attribute ($1$ less than for unranked data). Nevertheless, for a factor model with $6$ factors, clusters identical to those for unranked data were obtained (Figure \ref{breastAbs}).

\section{Conclusions}
Typically, any data to be analyzed is given in the form of tables containing m rows and n columns. Each column in the table represents the values of some attribute, and each row in the table represents one object in an n-dimensional attribute space. One of the issues of unsupervised machine learning is object clustering. This is an issue so common that many algorithms used for object clustering are known: the k-means method, density methods, spectral methods, hierarchical methods, etc. On the other hand, attribute clustering is not as obvious as object clustering, especially when the attributes are nominal.

An example of clustering numerical attributes is exploratory factor analysis, which aims to find subsets of attributes that are similar to factors. However, for exploratory factor analysis to work, the clustered attributes must take numeric values. For nominal attributes, it is not possible to calculate a correlation matrix, and therefore factor analysis is not possible either.

This article proposes a new attribute clustering algorithm also for the case where the clustered attributes do not include only numeric attributes, but also nominal attributes. The proposed algorithm also uses exploratory factor analysis. However, before the factor analysis starts, at the beginning of the algorithm's operation, the nominal data are encoded with numbers. 
When a given nominal attribute does not have equicardinal classes of identical elements, its values are encoded by the cardinalities of the classes. 
On the other hand, when an attribute has equicardinal classes of identical values, then the elements in these classes cannot be encoded with class cardinalities, because after encoding the individual classes would be indistinguishable. In such a situation, the one-hot encoding method is used to encode the nominal attribute.
The one-hot encoding method encodes an attribute containing $k$ classes using $k$ new binary attributes, each representing 1 class. After encoding, a factor analysis is performed   for the new set of numerical attributes, which models the encoded attributes using some smaller number of factors. The number of factors is chosen so that the factor model represents most of the variance of each attribute. After rotating the factors using the Varimax method, the encoded attributes are divided into different classes according to their similarity to the factors. An attribute is similar to a factor when the factor represents most of its variance. In this case, two similarity rules are used: the absolute majority of variance rule and the relative majority of variance rule.
In the case of the absolute majority variance rule, the common variance of an attribute and a factor is greater than $50\%$. In the case of the relative majority of variance rule, the common variance of an attribute and a factor may be less than $50\%$, but its level is greater than the attribute variance represented by the other factors. After the clustering procedure, non-trivial clusters, i.e., clusters that contain more than one attribute similar to a factor, are considered to be of interest.

The presented algorithm was used to cluster attributes from four different datasets. Because of the assumption made that all attributes in the \textit{Simple Weather Forecast} dataset are nominal attributes, the first clustering example was formal in nature. Its purpose was to explain in detail all the steps of the algorithm, from the numerical encoding of attributes, through the selection of the appropriate number of factors, to the final clustering of attributes. The remaining 3 datasets (\textit{Mushroom}, \textit{Automobile}, \textit{Breast Cancer}), were used to show that clusters of attributes similar to factors can be identified. 
Even a superficial analysis of directed graphs, which represent non-trivial clusters of attributes similar to factors, leads to the conclusion that the results obtained can be inspiring from the point of view of further inference about the relationships occurring between attributes.

In this article, such concepts as random component, standardization, correlation, common variance, etc. appeared. It should be noted that all these concepts are well interpretable only when the random attributes under analysis have a normal distribution. In other cases, such interpretation is difficult or impossible. When analyzing encoded nominal data with several classes of values, trying to interpret the above concepts is meaningless. On the other hand, the proposed algorithm offers meaningful and interpretable results. In order to be able to use them, the author of the article suggests not to interpret the above concepts in a statistical sense, but only to interpret them formally in a geometric sense. Accordingly, the random component is a vector, the standardization is a linear transformation that does not change the direction of the vector, the correlation is the cosine of the angle between two vectors, and the common variance is the square of this cosine.

Finally, a note on the possibility of using the algorithm proposed above to cluster (random) attributes in cases other than simple data types (numeric, nominal).
Here we focus on the applicability of this algorithm to the analysis of social research, which is conducted using surveys containing questions with answer choices. In the case of a question with a set of single-choice answers, it is possible to treat the obtained answers as nominal attribute values. When the obtained set of answers does not contain equicardinal classes, the answers should be encoded with their cardinality. When the obtained set of answers contains equicardinal classes, they should be encoded using the one-hot encoding method. On the other hand, for questions with multiple-choice answers, the one-hot encoding method is the only possible way to encode the answers.

\section*{Acknowledgments}
The author would like to express his gratitude to Dr. Waldemar Labuda for reading and reviewing this article before its publication.
The author would also like to acknowledge that he used support tools while writing this article. For translation, he mainly used the Deepl translation platform, available at \url{https://www.deepl.com}, and occasionally the Google Translate platform, available at \url{https://translate.google.com}. To create graphs representing the clusters, the author used the yEd application, available at \url{https://www.yworks.com/editors}.

\bibliography{attribCluster}\label{bibliography}

\begin{thebibliography}{10}
\providecommand{\url}[1]{#1}
\csname url@samestyle\endcsname
\providecommand{\newblock}{\relax}
\providecommand{\bibinfo}[2]{#2}
\providecommand{\BIBentrySTDinterwordspacing}{\spaceskip=0pt\relax}
\providecommand{\BIBentryALTinterwordstretchfactor}{4}
\providecommand{\BIBentryALTinterwordspacing}{\spaceskip=\fontdimen2\font plus
\BIBentryALTinterwordstretchfactor\fontdimen3\font minus
  \fontdimen4\font\relax}
\providecommand{\BIBforeignlanguage}[2]{{%
\expandafter\ifx\csname l@#1\endcsname\relax
\typeout{** WARNING: IEEEtran.bst: No hyphenation pattern has been}%
\typeout{** loaded for the language `#1'. Using the pattern for}%
\typeout{** the default language instead.}%
\else
\language=\csname l@#1\endcsname
\fi
#2}}
\providecommand{\BIBdecl}{\relax}
\BIBdecl

\bibitem{Stevens1946}
\BIBentryALTinterwordspacing
S.~S. Stevens, ``On the theory of scales of measurement,'' \emph{Science}, vol.
  103, no. 2684, pp. 677--680, 1946. [Online]. 
  \url{http://expsylab.psych.uoa.gr/fileadmin/expsylab.psych.uoa.gr/uploads/papers/Stevens_1946.pdf}
\BIBentrySTDinterwordspacing

\bibitem{Blalock1960}
\BIBentryALTinterwordspacing
H.~M. Blalock, \emph{{Social Statistics}}.\hskip 1em plus 0.5em minus
  0.4em\relax McGraw-Hill, 1960. [Online]. 
  \url{https://gwern.net/doc/statistics/1960-blalock-socialstatistics.pdf}
\BIBentrySTDinterwordspacing

\bibitem{Gniazdowski2015}
\BIBentryALTinterwordspacing
Z.~Gniazdowski and M.~Grabowski, ``{Numerical Coding of Nominal Data},''
  \emph{Zeszyty Naukowe WWSI}, vol.~9, no.~12, pp. 53--61, 2015. [Online].
  \url{https://www.doi.org/10.26348/znwwsi.12.53}
\BIBentrySTDinterwordspacing

\bibitem{Berry2018}
\BIBentryALTinterwordspacing
J.~E. J. K.~J. Berry, P.~W. Mielke~Jr, Berry, and Hiripi, \emph{Measurement of
  Association}.\hskip 1em plus 0.5em minus 0.4em\relax Springer, 2018.
  [Online]. 
  \url{https://link.springer.com/content/pdf/10.1007/978-3-319-98926-6.pdf}
\BIBentrySTDinterwordspacing

\bibitem{Hastie2017}
\BIBentryALTinterwordspacing
T.~Hastie, R.~Tibshirani, and J.~H. Friedman, \emph{The elements of statistical
  learning: data mining, inference, and prediction}.\hskip 1em plus 0.5em minus
  0.4em\relax Springer, 2017. [Online]. 
  \url{https://link.springer.com/content/pdf/10.1007/978-0-387-84858-7.pdf}
\BIBentrySTDinterwordspacing

\bibitem{Han2022}
J.~Han, J.~Pei, and H.~Tong, \emph{Data mining: concepts and techniques}.\hskip
  1em plus 0.5em minus 0.4em\relax Morgan Kaufmann, 2022.

\bibitem{Gniazdowski2022}
\BIBentryALTinterwordspacing
Z.~Gniazdowski, ``{On the Analysis of Correlation Between Nominal Data and
  Numerical Data},'' \emph{Zeszyty Naukowe WWSI}, vol.~16, no.~27, pp. 57--82,
  2022. [Online]. \url{https://www.doi.org/10.26348/znwwsi.27.57}
\BIBentrySTDinterwordspacing

\bibitem{Gniazdowski2013}
\BIBentryALTinterwordspacing
Z.~Gniazdowski, ``Geometric interpretation of a correlation,'' \emph{Zeszyty Naukowe
  WWSI}, vol.~7, no.~9, pp. 27--35, 2013. [Online]. 
  \url{https://www.doi.org/10.26348/znwwsi.9.27}
\BIBentrySTDinterwordspacing

\bibitem{Wilcoxon1992}
\BIBentryALTinterwordspacing
F.~Wilcoxon, \emph{Individual Comparisons by Ranking Methods}.\hskip 1em plus
  0.5em minus 0.4em\relax New York, NY: Springer New York, 1992, pp. 196--202.
  [Online]. \url{https://doi.org/10.1007/978-1-4612-4380-9_16}
\BIBentrySTDinterwordspacing

\bibitem{Gniazdowski2021}
\BIBentryALTinterwordspacing
Z.~Gniazdowski, ``{Principal Component Analysis versus Factor Analysis},''
  \emph{Zeszyty Naukowe WWSI}, vol.~15, no.~24, pp. 35--88, 2021. [Online].
  \url{https://www.doi.org/10.26348/znwwsi.24.35}
\BIBentrySTDinterwordspacing

\bibitem{Styan1973}
\BIBentryALTinterwordspacing
G.~P.~H. Styan, ``Hadamard products and multivariate statistical analysis,''
  \emph{Linear Algebra and its Applications}, vol.~6, pp. 217--240, 1973.
  [Online]. \url{https://doi.org/10.1016/0024-3795(73)90023-2}
\BIBentrySTDinterwordspacing

\bibitem{Kaiser1958}
\BIBentryALTinterwordspacing
H.~F. Kaiser, ``The varimax criterion for analytic rotation in factor
  analysis,'' \emph{Psychometrika}, vol.~23, no.~3, pp. 187--200, 1958.
  [Online]. \url{https://doi.org/10.1007/BF02289233}
\BIBentrySTDinterwordspacing

\bibitem{Majority2024}
\BIBentryALTinterwordspacing
``Majority,'' 2024. [Online]. 
  \url{https://en.wikipedia.org/wiki/Majority}
\BIBentrySTDinterwordspacing

\bibitem{Majority_rule2024}
\BIBentryALTinterwordspacing
``Majority rule,'' 2024. [Online]. 
  \url{https://en.wikipedia.org/wiki/Majority_rule}
\BIBentrySTDinterwordspacing

\bibitem{Plurality2024}
\BIBentryALTinterwordspacing
``Plurality (voting),'' 2024. [Online]. 
  \url{https://en.wikipedia.org/wiki/Plurality_(voting)}
\BIBentrySTDinterwordspacing

\bibitem{Pogonowski}
\BIBentryALTinterwordspacing
J.~Pogonowski, ``Własności relacji.'' [Online]. 
  \url{http://logic.amu.edu.pl/images/b/bb/Dygdwa.pdf}
\BIBentrySTDinterwordspacing

\bibitem{Pogonowski1997}
\BIBentryALTinterwordspacing
J.~Pogonowski, \emph{Przestrzenie Podobienstwa i Opozycji}.\hskip 1em plus 0.5em minus
  0.4em\relax Wydział Filozofii i Socjologii Uniwersytetu Warszawskiego, 1997,
  pp. 83--95. [Online]. 
  \url{https://logic.amu.edu.pl/images/8/85/Wolniew.pdf}
\BIBentrySTDinterwordspacing

\bibitem{Cerda2018}
\BIBentryALTinterwordspacing
P.~Cerda, G.~Varoquaux, and B.~K{\'e}gl, ``Similarity encoding for learning
  with dirty categorical variables,'' \emph{Machine Learning}, vol. 107, no.~8,
  pp. 1477--1494, 2018. [Online]. 
  \url{https://doi.org/10.1007/s10994-018-5724-2}
\BIBentrySTDinterwordspacing

\bibitem{Hancock2020}
\BIBentryALTinterwordspacing
J.~T. Hancock and T.~M. Khoshgoftaar, ``Survey on categorical data for neural
  networks,'' \emph{Journal of big data}, vol.~7, no.~1, p.~28, 2020. [Online].
  \url{https://doi.org/10.1186/s40537-020-00305-w}
\BIBentrySTDinterwordspacing

\bibitem{deepAIglossary}
\BIBentryALTinterwordspacing
``One hot encoding.'' [Online]. 
  \url{https://deepai.org/machine-learning-glossary-and-terms/one-hot-encoding}
\BIBentrySTDinterwordspacing

\bibitem{weather2022}
\BIBentryALTinterwordspacing
D.~Bhat, ``Simple weather forecast,'' 2022. [Online]. 
  \url{https://www.kaggle.com/datasets/dheemanthbhat/simple-weather-forecast}
\BIBentrySTDinterwordspacing

\bibitem{mushroom1981}
\BIBentryALTinterwordspacing
``Mushroom,'' UCI Machine Learning Repository, 1981. [Online]. 
  \url{https://doi.org/10.24432/C5959T}
\BIBentrySTDinterwordspacing

\bibitem{automobile1985}
\BIBentryALTinterwordspacing
J.~Schlimmer, ``{Automobile},'' UCI Machine Learning Repository, 1985.
  [Online]. \url{https://doi.org/10.24432/C5B01C}
\BIBentrySTDinterwordspacing

\bibitem{breastCancer1988}
\BIBentryALTinterwordspacing
M.~Zwitter and M.~Soklic, ``{Breast Cancer},'' UCI Machine Learning Repository,
  1988. [Online]. \url{https://doi.org/10.24432/C51P4M}
\BIBentrySTDinterwordspacing

\end{thebibliography}
\bibliographystyle{IEEEtran}

\end{document}